\documentclass{article}
\pdfoutput=1
\PassOptionsToPackage{numbers, compress}{natbib}

    \usepackage[final]{neurips_2022}

\usepackage[utf8]{inputenc} 
\usepackage[T1]{fontenc}    
\usepackage{url}            
\usepackage{booktabs}       
\usepackage{amsfonts}       
\usepackage{nicefrac}       
\usepackage{microtype}      
\usepackage{xcolor}         

\usepackage{import}
\usepackage{wrapfig}
\usepackage{graphicx}
\usepackage{subcaption}

\usepackage{amssymb}
\usepackage{amsthm}
\usepackage{mathtools}
\usepackage{bm}
\usepackage{xcolor}
\usepackage{accents}
\usepackage{cancel}
\usepackage{multicol}
\usepackage{algorithmicx}
\usepackage{algorithm}
\usepackage[noend]{algpseudocode}

\newtheorem{proposition}{Proposition}
\newtheorem{theorem}{Theorem}

\newtheorem{lemma}{Lemma}
\newtheorem{definition}{Definition}

\theoremstyle{remark}

\newcommand\indep{\protect\mathpalette{\protect\independenT}{\perp}}
\def\independenT#1#2{\mathrel{\rlap{$#1#2$}\mkern2mu{#1#2}}}

\DeclareMathOperator*{\argmin}{arg\,min}

\newcommand{\E}{\mathop{\mathbb{E}}}

\newcommand{\D}{\mathcal{D}}

\newcommand{\Xcal}{\mathcal{X}}
\newcommand{\X}{\mathbf{X}}
\newcommand{\x}{\mathbf{x}}

\newcommand{\hu}{\mathrm{u}}

\newcommand{\Tcal}{\mathcal{T}}
\newcommand{\Tf}{\mathrm{T}}
\newcommand{\tf}{\mathrm{t}}
\newcommand{\tfp}{\tf^{\prime}}

\newcommand{\Ycal}{\mathcal{Y}}
\newcommand{\Y}{\mathrm{Y}}
\newcommand{\y}{\mathrm{y}}
\newcommand{\ystar}{\y^{*}}
\newcommand{\yH}{\y_{H}}

\newcommand{\Yt}{\mathrm{Y}_{\tf}}
\newcommand{\YT}{\mathrm{Y}_{\Tf}}
\newcommand{\yt}{\mathrm{y}_{\tf}}

\newcommand{\mut}{\widetilde{\mu}}

\newcommand{\sigmat}{\widetilde{\sigma}}

\newcommand{\pitilde}{\widetilde{\pi}}

\newcommand{\WndH}{\mathcal{W}_{\textrm{nd}}^{H}}
\newcommand{\WniH}{\mathcal{W}_{\textrm{ni}}^{H}}

\newcommand{\params}{\bm{\theta}}
\newcommand{\Params}{\bm{\Theta}}

\newcommand{\CI}{\mathrm{CI}}

\definecolor{mydarkblue}{rgb}{0,0.08,0.45}
\definecolor{purple}{HTML}{9673A6}
\definecolor{red}{HTML}{B85450}
\definecolor{blue}{HTML}{6C8EBF}
\definecolor{orange}{HTML}{D79B00}
\definecolor{green}{HTML}{82B366}
\usepackage{hyperref}       
\hypersetup{
    colorlinks=True,
    linkcolor=mydarkblue,
    citecolor=mydarkblue,
    filecolor=mydarkblue,
    urlcolor=mydarkblue,
}
\usepackage{cleveref}

\title{Scalable Sensitivity and Uncertainty Analyses for Causal-Effect Estimates of Continuous-Valued Interventions}

\author{%
    Andrew Jesson\thanks{Correspondence to \texttt{andrew.jesson@cs.ox.ac.uk}}\\
    OATML \\ 
    Department of Computer Science \\ 
    University of Oxford \\
    \And
    Alyson Douglas \\
    AOPP \\ 
    Department of Physics \\
    University of Oxford \\
    \And
    Peter Manshausen \\
    AOPP \\ 
    Department of Physics \\
    University of Oxford \\
    \And
    Maëlys Solal \\
    Department of Computer Science \\
    University of Oxford \\
    \And
    Nicolai Meinshausen \\
    Seminar for Statistics \\ 
    Department of Mathematics \\ 
    ETH Zurich \\
    \And
    Philip Stier \\
    AOPP \\ 
    Department of Physics \\
    University of Oxford \\
    \And
    Yarin Gal \\
    OATML \\ 
    Department of Computer Science \\ 
    University of Oxford \\
    \And
    Uri Shalit \\
    Machine Learning and Causal Inference in Healthcare Lab \\ 
    Technion -- Israel Institute of Technology  \\
}

\begin{document}

\maketitle

\begin{abstract}
    Estimating the effects of continuous-valued interventions from observational data is a critically important task for climate science, healthcare, and economics. 
    Recent work focuses on designing neural network architectures and regularization functions to allow for scalable estimation of average and individual-level dose-response curves from high-dimensional, large-sample data. 
    Such methodologies assume ignorability (observation of all confounding variables) and positivity (observation of all treatment levels for every covariate value describing a set of units), assumptions problematic in the continuous treatment regime. 
    Scalable sensitivity and uncertainty analyses to understand the ignorance induced in causal estimates when these assumptions are relaxed are less studied.
    Here, we develop a continuous treatment-effect marginal sensitivity model (CMSM) and derive bounds that agree with the observed data and a researcher-defined level of hidden confounding. 
    We introduce a scalable algorithm and uncertainty-aware deep models to derive and estimate these bounds for high-dimensional, large-sample observational data.
    We work in concert with climate scientists interested in the climatological impacts of human emissions on cloud properties using satellite observations from the past 15 years. 
    This problem is known to be complicated by many unobserved confounders.
\end{abstract}

\section{Introduction}
\label{sec:introduction}
Understanding the causal effect of a continuous variable (termed ``treatment'') on individual units and subgroups is crucial across many fields. 
In economics, we might like to know the effect of price on demand from different customer demographics. 
In healthcare, we might like to know the effect of medication dosage on health outcomes for patients of various ages and comorbidities. 
And in climate science, we might like to know the effects of anthropogenic emissions on cloud formation and lifetimes under variable atmospheric conditions. 
In many cases, these effects must be estimated from observational data as experiments are often costly, unethical, or otherwise impossible to conduct. 

Estimating causal effects from observational data can only be done under certain conditions, some of which are not testable from data. 
The most prominent are the common assumptions that all confounders between treatment and outcome are measured (``no hidden confounders''), and any level of treatment could occur for any observable covariate vector (``positivity''). 
These assumptions and their possible violations introduce uncertainty when estimating treatment effects. 
Estimating this uncertainty is crucial for decision-making and scientific understanding.
For example, understanding how unmeasured confounding can change estimates about the impact of emissions on cloud properties can help to modify global warming projection models to account for the uncertainty it induces.  

We present a novel marginal sensitivity model for continuous treatment effects. 
This model is used to develop a method that gives the user a corresponding interval representing the ``ignorance region'' of the possible treatment outcomes per covariate and treatment level \citep{d2019multi} for a specified level of violation of the no-hidden confounding assumption. 
We adapt prior work \citep{tan2006msm,kallus2019interval,jesson2021quantifying} to the technical challenge presented by continuous treatments. 
Specifically, we modify the existing model to work with propensity score densities instead of propensity score probabilities (see \Cref{sec:cmsm} below) and propose a method to relate ignorability violations to the unexplained range of outcomes.
Further, we derive bootstrapped uncertainty intervals for the estimated ignorance regions and show how to efficiently compute the intervals, thus providing a method for quantifying the uncertainty presented by finite data and possible violations of the positivity assumption. 
We validate our methods on synthetic data and provide an application on real-world satellite observations of the effects of anthropogenic emissions on cloud properties. 
For this application, we develop a new neural network architecture for estimating continuous treatment effects that can take into account spatiotemporal covariates. 
We find that the model accurately captures known patterns of cloud deepening in response to anthropogenic emission loading with realistic intervals of uncertainty due to unmodeled confounders in the satellite data.
\section{Problem Setting}
Let the random variable $\X \in \Xcal$ model observable covariates.
For clarity, we will assume that $\Xcal$ is a $d$-dimensional continuous space: $\Xcal \subseteq \mathbb{R}^d$, but this does not preclude more diverse spaces.
Instances of $\X$ are denoted by $\x$.
The observable continuous treatment variable is modeled as the random variable $\Tf \in \Tcal \subseteq \mathbb{R}$.
Instances of $\Tf$ are denoted by $\tf$.
Let the random variable $\Y \in \Ycal \subseteq \mathbb{R}$ model the observable continuous outcome variable.
Instances of $\Y$ are denoted by $\y$.
Using the Neyman-Rubin potential outcomes framework \citep{neyman1923edited, rubin1974estimating, sekhon2008neyman}, we model the potential outcome of a treatment level $\tf$ by the random variable $\Yt \in \Ycal$.
Instances of $\Yt$ are denoted by $\yt$.
We assume that the observational data, $\mathcal{D}_n$, consists of $n$ realizations of the random variables, $\mathcal{D}_n = \left\{ (\x_i, \tf_i, \y_i) \right\}_{i=1}^{n}$.
We let the observed outcome be the potential outcome of the assigned treatment level, $\y_i = \y_{\tf_i}$, thus assuming non-interference and consistency \citep{rubin1980randomization}.
Moreover, we assume that the tuple $(\x_i, \tf_i, \y_i)$ are i.i.d. samples from the joint distribution $P(\X, \Tf, \Y_\Tf)$, where $\YT = \{\Yt: \tf \in \Tcal\}$.

We are interested in the \textbf{conditional average potential outcome (CAPO)} function, $\mu(\x, \tf)$, and the \textbf{average potential outcome (APO)} — or dose-response function — $\mu(\tf)$, for continuous valued treatments.
These functions are defined by the expectations:
\begin{minipage}{.5\linewidth}
    \begin{equation}
        \label{eq:CAPO}
        \mu(\x, \tf) \coloneqq \E\left[ \Yt \mid \X = \x \right]
    \end{equation}
\end{minipage}%
\begin{minipage}{.5\linewidth}
    \begin{equation}
        \label{eq:APO}
        \mu(\tf) \coloneqq \E\left[ \mu(\X, \tf) \right].
    \end{equation}
\end{minipage}

Under the assumptions of ignorability, $\YT \indep \Tf \mid \X$, and positivity, $p(\tf \mid \X = \x) > 0 : \forall \tf \in \Tcal, \forall \x \in \Xcal$ — jointly  known as \emph{strong ignorability}  \citep{rosenbaum1983central} — the CAPO and APO are identifiable from the observational distribution $P(\X, \Tf, \Y_\Tf)$ as:
\noindent
\begin{minipage}{.5\linewidth}
    \begin{equation}
        \label{eq:CAPO_stat}
         \mut(\x, \tf) = \E\left[ \Y \mid \Tf = \tf, \X = \x \right]
    \end{equation}
\end{minipage}%
\begin{minipage}{.5\linewidth}
    \begin{equation}
        \label{eq:APO_stat}
         \mut(\tf) = \E\left[ \mut(\X, \tf) \right].
    \end{equation}
\end{minipage}

In practice, however, these assumptions rarely hold. For example, there will almost always be unobserved confounding variables, thus violating the ignorability (also known as unconfoundedness or exogeneity) assumption, $\YT \not\indep \Tf \mid \X$.
Moreover, due to both the finite sample of observed data, $\D$, and also the continuity of treatment $\Tf$, there will most certainly be values, $\Tf=\tf$, that are unobserved for a given covariate measurement, $\X=\x$, leading to violations or near violations of the positivity assumption (also known as overlap).
\section{Methods}
\label{sec:methods}

\label{sec:cmsm}
We propose the continuous marginal sensitivity model (CMSM) as a new marginal sensitivity model (MSM \citep{tan2006msm}) for continuous treatment variables.
The set of conditional distributions of the potential outcomes given the observed treatment assigned, 
$\left\{ P(\Yt \mid \Tf=\tf, \X=\x) : \tf \in \Tcal \right\}$,
are identifiable from data, $\D$. 
But, the set of marginal distributions of the potential outcomes, $\left\{ P(\Yt \mid, \X=\x) : \tf \in \Tcal \right\}$, each given as a continuous mixture,
\begin{equation*}
   P(\Yt \mid \X=\x) = \int_{\Tcal} p(\tfp \mid \x) P(\Yt \mid \Tf=\tfp, \X=\x) d\tfp,
\end{equation*}
are not.
This is due to the general unidentifiability of the component distributions, $P(\Yt \mid \Tf=\tfp, \X=\x)$, where $\Yt$ cannot be observed for units under treatment level $\Tf = \tfp$ for $\tfp \neq \tf$: the well-known ``fundamental problem of causal inference'' \citep{holland1986statistics}.
Yet, under the ignorability assumption, the factual $P(\Yt \mid \Tf=\tf, \X=\x)$ and counterfactual $P(\Yt \mid \Tf=\tfp, \X=\x)$ are equal for all $\tfp \in \Tcal$.
Thus, $P(\Yt \mid \X=\x)$ and $P(\Yt \mid \Tf=\tf, \X=\x)$ are identical, and any divergence between them is indicative of hidden confounding. 
But, such divergence is not observable in practice.

The CMSM supposes a degree of divergence between the unidentifiable $P(\Yt \mid \X=\x)$ and the identifiable $P(\Yt \mid \Tf=\tf, \X=\x)$ by assuming that the rate of change of $P(\Yt \mid \X=\x)$ with respect to $P(\Yt \mid \Tf=\tf, \X=\x)$ is bounded by some value greater than or equal to $1$.
The Radon-Nikodym derivative formulates the divergence,
$
    \lambda(\yt; \x, \tf) = \frac{dP(\Yt \mid \X=\x)}{dP(\Yt \mid \Tf=\tf, \X=\x)},
$
under the assumption that $P(\Yt \mid \X=\x)$ is absolutely continuous with respect to $P(\Yt \mid \Tf=\tf, \X=\x)$, $\forall\tf \in \Tcal$.
\begin{proposition}
    \label{prop:density_ratio}
    Under the additional assumption that $P(\Yt \mid \Tf=\tf, \X=\x)$ and the Lebesgue measure are mutually absolutely continuous, the Radon-Nikodym derivative above is equal to the ratio between the unidentifiable ``complete'' propensity density for treatment $p(\tf \mid \yt, \x)$ and the identifiable ``nominal'' propensity density for treatment $p(\tf \mid \x)$,
    \begin{equation}
        \lambda(\yt; \x, \tf) = \frac{p(\tf \mid \x)}{p(\tf \mid \yt, \x)},
    \end{equation}
    Proof (\Cref{proof:density_ratio}) and an analysis of this proposition are given in \Cref{app:cmsm}.
\end{proposition}

The value $\lambda(\yt; \x, \tf)$ cannot be identified from the observational data alone; the merit of the CMSM is that enables a domain expert to express their belief in what is a plausible degree hidden confounding through the parameter $\Lambda \geq 1$. Where, $\Lambda^{-1} \leq p(\tf \mid \x) / p(\tf \mid \yt, \x) \leq \Lambda$,
reflects a hypothesis that the ``complete'', unidentifiable propensity density for subjects with covariates $\X=\x$ can be different from the identifiable ``nominal'' propensity density by at most a factor of $\Lambda$.
These inequalities allow for the specification of user hypothesized complete propensity density functions, $p(\tf \mid \y, \x)$, and we define the CMSM as the set of such functions that agree with the inequalities.
\begin{definition}
    Continuous Marginal Sensitivity Model (CMSM)
    \begin{equation}
        \mathcal{P}(\Lambda) \coloneqq \left\{ p(\tf \mid \y, \x): \frac{1}{\Lambda} \leq \frac{p(\tf \mid \x)}{p(\tf \mid \yt, \x)} \leq \Lambda, \forall \y \in \mathbb{R}, \forall \x \in \Xcal \right\}
    \end{equation}
\end{definition}

\textbf{Remark.} Note that the CMSM is defined in terms of a \emph{density ratio}, $p(\tf \mid \x) / p(\tf \mid \yt, \x)$, whereas the MSM for binary-valued treatments is defined in terms of an \emph{odds ratio}, $\frac{P(\tf \mid \x)}{(1 - P(\tf \mid \x))} / \frac{P(\tf \mid \yt, \x)}{(1 - P(\tf \mid \yt, \x))}$. 
Importantly, naively substituting densities into the MSM for binary-treatments would violate the condition that $\lambda>0$ as the densities $p(\tf \mid \x)$ or $p(\tf \mid \yt, \x)$ can each be greater than one, which would result in a negative $1 - p(\tf \mid \cdot)$.
The odds ratio is familiar to practitioners. 
The density ratio is less so.
We offer a transformation of the sensitivity analysis parameter $\Lambda$ in terms of the unexplained range of the outcome later.
\subsection{Continuous Treatment Effect Bounds Without Ignorability}
\label{sec:bounds}
The CAPO and APO (dose-response) functions cannot be point identified from observational data without ignorability.
Under the CMSM with a given $\Lambda$, we can only identify a set of CAPO and APO functions jointly consistent with the observational data $\D$ and the continuous marginal sensitivity model.
All of the functions in this set are possible from the point of view of the observational data alone. 
So to cover the range of all possible functional values, we seek an interval function that maps covariate values, $\X=\x$, to the upper and lower bounds of this set for every treatment value, $\tf$.

For $\tf \in \Tcal$ and $\x \in \Xcal$, let $p(\yt \mid \tf, \x)$ denote the density of the distribution $P(\Yt \mid \Tf=\tf, \X=\x)$.
As a reminder, this distribution is identifiable from observational data, but without further assumptions the CAPO, $\mu(\x, \tf) = \E\left[ \Yt \mid \X = \x \right]$, is not.
We can express the CAPO in terms of its identifiable and unidentifiable components as
\begin{equation}
    \begin{split}
        \mu(\x, \tf)
        &= \frac{\int_\Ycal \yt \frac{p(\yt \mid \tf, \x)}{p(\tf \mid \yt, \x)}d\yt}{\int_\Ycal \frac{p(\yt \mid \tf, \x)}{p(\tf \mid \yt, \x)} d\yt}
        = \mut(\x, \tf) + \frac{\int_\Ycal w(\y, \x) (\y - \mut(\x, \tf)) p(\y \mid \tf, \x)d\y}{(\Lambda^2 - 1)^{-1} + \int_\Ycal w(\y, \x)p(\y \mid \tf, \x)d\y}, \\
        &\equiv \mu(w(\y, \x); \x, \tf, \Lambda)
    \end{split}
    \label{eq:mu_w}
\end{equation}
where, by a one-to-one change of variables, $\frac{1}{p(\tf \mid \yt, \x)} = \frac{1}{\Lambda p(\tf \mid \x)} + w(\y, \x) (\frac{\Lambda}{p(\tf \mid \x)} - \frac{1}{\Lambda p(\tf \mid \x)})$ with $w : \Ycal\times \Xcal \to [0, 1]$.
Both \cite{kallus2019interval} and later \cite{jesson2021quantifying} provide analogous expressions for the CAPO in the discrete treatment regime under the MSM, and we provide our derivation in \Cref{lem:capo}. 

The uncertainty set that includes all possible values of $w(\y, \x)$ that agree with the CMSM, \emph{i.e.}, the set of functions that violate ignorability by no more than $\Lambda$, can now be expressed as 
$
    \mathcal{W} = \{w: w(\y, \x) \in [ 0, 1 ] \quad \forall \y \in \Ycal,  \forall \x \in \Xcal \}.
$

With this set of functions, we can now define the CAPO and APO bounds under the CMSM.
The CAPO lower, $\underline{\mu}(\x, \tf; \Lambda)$, and upper, $\overline{\mu}(\x, \tf; \Lambda)$, bounds under the CMSM with parameter $\Lambda$ are:
\noindent
\begin{minipage}{.5\linewidth}
    \begin{equation}
        \label{eq:CAPO_lower}
        \begin{split}
            \underline{\mu}(\x, \tf; \Lambda) 
            &\coloneqq \inf_{w \in \mathcal{W}} \mu(w(\y, \x); \x, \tf, \Lambda) \\
            &= \inf_{w \in \WniH} \mu(w(\y); \x, \tf, \Lambda) 
        \end{split}
    \end{equation}
\end{minipage}%
\begin{minipage}{.5\linewidth}
    \begin{equation}
        \label{eq:CAPO_upper}
        \begin{split}
            \overline{\mu}(\x, \tf; \Lambda) 
            &\coloneqq \sup_{w \in \mathcal{W}} \mu(w(\y, \x); \x, \tf, \Lambda) \\
            &= \sup_{w \in \WndH} \mu(w(\y); \x, \tf, \Lambda)
        \end{split}
    \end{equation}
\end{minipage}
\break
Where the sets $\WniH = \left\{ w: w(y) = H(\yH - \y) \right\}_{\yH \in \Ycal}$, and $\WndH = \left\{ w: w(y) = H(\y - \yH)  \right\}_{\yH \in \Ycal}$, and $H(\cdot)$ is the Heaviside step function. 
\Cref{lem:monotonic} in \cref{app:step} proves the equivalence in \cref{eq:CAPO_upper} for bounded $Y$. 
The equivalence in \cref{eq:CAPO_lower} can be proved analogously.

The APO lower, $\underline{\mu}(\tf; \Lambda)$, and upper, $\overline{\mu}(\tf; \Lambda)$, bounds under the CMSM with parameter $\Lambda$ are:
\noindent
\begin{minipage}{.5\linewidth}
    \begin{equation}
        \label{eq:APO_lower}
        \underline{\mu}(\tf; \Lambda) \coloneqq \E\left[ \underline{\mu}(\X, \tf; \Lambda) \right]
    \end{equation}
\end{minipage}%
\begin{minipage}{.5\linewidth}
    \begin{equation}
        \label{eq:APO_upper}
        \overline{\mu}(\tf; \Lambda) \coloneqq \E\left[ \overline{\mu}(\X, \tf; \Lambda) \right]
    \end{equation}
\end{minipage}

\textbf{Remark.} It is worth pausing here and breaking down \Cref{eq:mu_w} to get an intuitive sense of how the specification of $\Lambda$ in the CMSM affects the bounds on the causal estimands. 
When $\Lambda \to 1$, then the $(\Lambda^2 - 1)^{-1}$ term (and thus the denominator) in \Cref{eq:mu_w} tends to infinity.
As a result, the CAPO under $\Lambda$ converges to the empirical estimate of the CAPO — $\mu(w(\y); \x, \tf, \Lambda \to 1) \to \mut(\x, \tf)$ — as expected.
Thus, the supremum and infimum in \Cref{eq:CAPO_lower,eq:CAPO_upper} become independent of $w$, and the ignorance intervals concentrate on point estimates.
Next, consider complete relaxation of the ignorability assumption, $\Lambda \to \infty$. Then, the $(\Lambda^2 - 1)^{-1}$ term tends to zero, and we are left with,
\begin{equation*}
    \begin{split}
        \mu(w; \cdot, \Lambda \to \infty)
        \to \mut(\x, \tf) + \frac{\int_\Ycal w(\y) (\y - \mut(\x, \tf)) p(\y \mid \tf, \x)d\y}{\int_\Ycal w(\y)p(\y \mid \tf, \x)d\y,}
        = \mut(\x, \tf) + \!\!\!\! \E_{p(w(\y) \mid \x, \tf)} \!\!\!\! [\Y - \mut(\x, \tf)],
    \end{split}
\end{equation*}
where, $p(w(\y) \mid \x, \tf) \equiv \frac{w(\y) p(\y \mid \tf, \x)}{\int_\Ycal w(\y')p(\y' \mid \tf, \x)d\y'}$, a distribution over $\Y$ given $\X=x$ and $\Tf=\tf$.
Thus, when we \emph{relax} the ignorability assumption entirely, the CAPO can be anywhere in the range of $\Y$.

The parameter $\Lambda$ relates to the proportion of unexplained range in $\Y$ assumed to come from unobserved confounders after observing $\x$ and $\tf$. 
When a user sets $\Lambda$ to 1, they assume that the entire unexplained range of $\Y$ comes from unknown mechanisms independent of $\Tf$. 
As the user increases $\Lambda$, they attribute some of the unexplained range of $\Y$ to mechanisms causally connected to $\Tf$.
For bounded $\Yt$, this proportion can be calculated as:
\begin{equation*}
        \rho(\x, \tf; \Lambda) \coloneqq \frac{\overline{\mu}(\x, \tf; \Lambda) - \underline{\mu}(\x, \tf; \Lambda)}{\overline{\mu}(\x, \tf; \Lambda \to \infty) - \underline{\mu}(\x, \tf; \Lambda \to \infty)}
        = \frac{\overline{\mu}(\x, \tf; \Lambda) - \underline{\mu}(\x, \tf; \Lambda)}{\y_{\max} - \y_{\min} \mid \X=x, \Tf=\tf}.
\end{equation*}
The user can sweep over a set of $\Lambda$ values and report the bounds corresponding to a $\rho$ value they deem tolerable (e.g., $\rho=0.5$ yields bounds for the assumption that half the unexplained range in $Y$ is due to unobserved confounders). For unbounded outcomes, the limits can be estimated empirically by increasing $\Lambda$ to a large value. Refer to \Cref{fig:rho} in the appendix for a comparison between $\rho$ and $\Lambda$.

For another way to interpret $\Lambda$, in \Cref{app:kld} we $\Lambda$ can be presented as a bound on the Kullback–Leibler divergence between the nominal and complete propensity scores  through the relationship: $|\log{\left(\Lambda\right)}| \geq  D_{\mathrm{KL}}\left( P(\Yt \mid \Tf=\tf, \X=\x) || P(\Yt \mid \X=\x) \right)$.
\subsection{Semi-Parametric Interval Estimator}
\label{sec:estimator}
\begin{wrapfigure}{R}{0.46\textwidth}
    \begin{minipage}{0.46\textwidth}
        \vspace{-2em}
        \begin{algorithm}[H]
            \caption{Grid Search Interval Optimizer}
            \label{alg:grid-search}
            \begin{algorithmic}[1]
                \Require{$\x$ is an instance of $\X$, $\tf$ is a treatment level to evaluate, $\Lambda$ is a belief in the amount of hidden confounding, $\params$ are optimized model parameters, $\widehat{\Ycal}$ is a set of unique values $\{\y \sim p(\y \mid \tf, \x, \params)\}$.}
                \Function{GridSearch}{$\x$, $\tf$, $\Lambda$, $\params$, $\widehat{\Ycal}$}
                    \State $\overline{\mu} \gets -\infty$, $\overline{\y} \gets 0$ 
                    \State $\underline{\mu} \gets \infty$, $\underline{\y} \gets 0$
                    \For{$\yH \in \widehat{\Ycal}$}
                        \State $\overline{\kappa} \gets \mu(H(\y - \yH); \x, \tf, \Lambda, \params)$
                        \State $\underline{\kappa} \gets \mu(H(\yH - \y); \x, \tf, \Lambda, \params)$
                        \If{$\overline{\kappa} > \overline{\mu}$}
                            \State $\overline{\mu}\gets \overline{\kappa}$, $\overline{\y} \gets \yH$
                        \EndIf
                        \If{$\underline{\kappa} < \underline{\mu}$}
                            \State $\underline{\mu} \gets \underline{\kappa}$, $\underline{\y} \gets \yH$
                        \EndIf
                    \EndFor
                    \State \Return $\underline{\y}, \overline{\y}$
                \EndFunction
            \end{algorithmic}
        \end{algorithm}
    \end{minipage}
\end{wrapfigure}
Following \cite{jesson2021quantifying}, we develop a semi-parametric estimator of the bounds in \cref{eq:CAPO_lower,eq:CAPO_upper,eq:APO_lower,eq:APO_upper}. 
Under assumption $\Lambda$, the bounds on the expected potential outcome over $\mu(w(\y); \x, \tf, \Lambda)$ are completely defined in terms of identifiable quantities: namely, the conditional density of the outcome given the assigned treatment and measured covariates, $p(\y \mid \tf, \x)$; and the conditional expected outcome $\mut(\x, \tf)$. 
Thus, we define a density estimator, $p(\y \mid \tf, \x, \params)$, and estimator, $\mu(\x, \tf; \params)$, parameterized by instances $\params$ of the random variable $\Params$.
The choice of density estimator is ultimately up to the user and will depend on the scale of the problem examined and the distribution of the outcome variable $\Y$. 
In \Cref{sec:scalable}, we will outline how to define appropriate density estimators for high-dimensional, large-sample, continuous-valued treatment problems.
Next, we need an estimator of the integrals in $\mu(w(\y); \x, \tf, \Lambda, \params)$, \cref{eq:mu_w}. 
We use Monte-Carlo (MC) integration to estimate the expectation of arbitrary functions $h(y)$ with respect to the parametric density estimate $p(\y \mid \tf, \x, \params)$:
$
    I(h(\y)) \coloneqq \frac{1}{m} \sum_{i=1}^m h(\y_i), \quad \y_i \sim p(\y \mid \tf, \x, \params).
$
We outline how the Gauss-Hermite quadrature rule is an alternate estimator of these expectations in \Cref{app:gauss-hermite}. The integral estimators allow for the semi-parametric estimators for the CAPO and APO bounds under the CMSM to be defined.

The semi-parametric CAPO bound estimators under the CMSM with sensitivity parameter $\Lambda$ are:
\noindent
\begin{minipage}{.5\linewidth}
    \begin{equation}
        \label{eq:CAPO_lower_param}
        \underline{\mu}(\x, \tf; \Lambda, \params) \coloneqq \inf_{w \in \WniH} \mu(w(\y); \x, \tf, \Lambda, \params)
    \end{equation}
\end{minipage}%
\begin{minipage}{.5\linewidth}
    \begin{equation}
        \label{eq:CAPO_upper_param}
        \overline{\mu}(\x, \tf; \Lambda, \params) \coloneqq \sup_{w \in \WndH} \mu(w(\y); \x, \tf, \Lambda, \params)
    \end{equation}
\end{minipage}
where,
\begin{equation*}
    \mu(w(\y); \x, \tf, \Lambda, \params) 
    \equiv \mut(\x, \tf; \params) + \frac{I\left( w(\y) (\y - \mut(\x, \tf; \params)) \right)}{(\Lambda^2 - 1)^{-1} + I\left( w(\y) \right)}.
\end{equation*}

The semi-parametric APO bound estimators under the CMSM with sensitivity parameter $\Lambda$ are:
\noindent
\begin{minipage}{.5\linewidth}
    \begin{equation}
        \label{eq:APO_lower_param}
        \underline{\mu}(\tf; \Lambda, \params) \coloneqq \E\left[ \underline{\mu}(\X, \tf; \Lambda, \params) \right]
    \end{equation}
\end{minipage}%
\begin{minipage}{.5\linewidth}
    \begin{equation}
        \label{eq:APO_upper_param}
        \overline{\mu}(\tf; \Lambda, \params) \coloneqq \E\left[ \overline{\mu}(\X, \tf; \Lambda, \params) \right]
    \end{equation}
\end{minipage}
\begin{theorem}
    \label{th:sharpness}
    In the limit of data ($n \to \infty$) and MC samples ($m \to \infty$), for observed $(\X=\x, \Tf=\tf) \in \D_n$, we assume that $p(\y \mid \tf, \x, \params)$ converges in measure to $p(\y \mid \tf, \x)$, $\mut(\x, \tf; \params)$ is a consistent estimator of $\mut(\x, \tf)$, and $p(\tf \mid \yt, \x)$ is bounded away from 0 uniformly for all $\yt \in \Ycal$. Then, $\underline{\mu}(\x, \tf; \Lambda, \params) \overset{p}{\to} \underline{\mu}(\x, \tf; \Lambda)$ and $\overline{\mu}(\x, \tf; \Lambda, \params) \overset{p}{\to} \overline{\mu}(\x, \tf; \Lambda)$. Proof in \Cref{app:sharpness}.
\end{theorem}

\subsection{Solving for w}
\label{sec:algorithm}
We are interested in a scalable algorithm to compute the intervals on the CAPO function, \cref{eq:CAPO_lower_param,eq:CAPO_upper_param}, and the APO (dose-response) function, \cref{eq:APO_lower_param,eq:APO_upper_param}.
The need for scalability stems not only from dataset size.
The intervals also need to be evaluated for arbitrarily many values of the continuous treatment variable, $\tf$, and the sensitivity parameter $\Lambda$.
The bounds on the CAPO function can be calculated independently for each instance $\x$, 
and the limits on the APO are an expectation over the CAPO function bounds.

The upper and lower bounds of the CAPO function under treatment, $\tf$, and sensitivity parameter, $\Lambda$, can be estimated for any observed covariate value, $\x$, as
\begin{equation*}
    \widehat{\underline{\mu}}(\x, \tf; \Lambda, \params) \coloneqq \mu(H(\underline{\y} - \y); \x, \tf, \Lambda, \params),
\end{equation*}
\begin{equation*}
    \widehat{\overline{\mu}}(\x, \tf; \Lambda, \params) \coloneqq \mu(H(\y - \overline{\y}); \x, \tf, \Lambda, \params),
\end{equation*}
where $\underline{\y}$ and $\overline{\y}$ are found using \Cref{alg:grid-search}.
See \Cref{alg:line-search} and \Cref{app:gradient} for optional methods.

The upper and lower bounds for the APO (dose-response) function under treatment $\Tf=\tf$ and sensitivity parameter $\Lambda$ can be estimated over any set of observed covariates $\D_\x = \{\x_i\}_{i=1}^n$, as
\noindent
\begin{minipage}{.45\linewidth}
    \begin{equation*}
        \widehat{\underline{\mu}}(\tf; \Lambda, \params) \coloneqq \frac{1}{n} \sum_{i = 1}^{n} \widehat{\underline{\mu}}(\x_i, \tf; \Lambda, \params),
    \end{equation*}
\end{minipage}%
\begin{minipage}{.45\linewidth}
    \begin{equation*}
        \widehat{\overline{\mu}}(\tf; \Lambda, \params) \coloneqq \frac{1}{n} \sum_{i = 1}^{n} \widehat{\overline{\mu}}(\x_i, \tf; \Lambda, \params),
    \end{equation*}
\end{minipage}%
\begin{minipage}{.1\linewidth}
    \begin{equation*}
        \x_i \in \D_\x.
    \end{equation*}
\end{minipage}

\subsection{Uncertainty about the Continuous Treatment Effect Interval}
\label{sec:uncertainty}
Following \cite{zhao2019sensitivity}, \cite{dorn2021sharp}, and \cite{chernozhukov2021omitted}, we construct $(1 - \alpha)$ statistical confidence intervals for the upper and lower bounds under the CMSM using the percentile bootstrap estimator.
\cite{jesson2020identifying} and \cite{jesson2021quantifying} have shown that statistical uncertainty is  appropriately high for regions with poor overlap.
Let $P_{\D}$ be the empirical distribution of the observed data sample, $\D=\{\x_i, \tf_i, \y_i\}_{i=1}^n = \{\mathbf{S}_i\}_{i=1}^n$.
Let $\widehat{P}_{\D} = \{\widehat{\D}_k\}_{k=1}^{n_b}$ be the bootstrap distribution over $n_b$ datasets, $\widehat{\D}_k=\{\widehat{\mathbf{S}}_i\}_{i=1}^n$, sampled with replacement from the empirical distribution, $P_{\D}$.
Let $Q_{\alpha}$ be the $\alpha$-quantile of $\mu(w(\y); \x, \tf, \Lambda, \params)$ in the bootstrap resampling distribution: 
$
Q_{\alpha} \coloneqq \inf_{\mu^*} \left\{\widehat{P}_\D(\mu(w(\y); \x, \tf, \Lambda, \params) \leq \mu^*) \geq \alpha \right\}.
$
Finally, let $\params_k$ be the parameters of the model of the $k$-th bootstrap sample of the data.
Then, the bootstrap confidence interval of the upper and lower bounds of the CAPO function under the CMSM is given by:
$
    \CI_b\left(\mu(\x, \tf; \Lambda, \alpha) \right) \coloneqq \left[ \underline{\mu}_{b}(\x, \tf; \Lambda, \alpha), \overline{\mu}_{b}(\x, \tf; \Lambda, \alpha) \right],
$
where,
\noindent
\begin{minipage}{.50\linewidth}
    \begin{equation*}
        \underline{\mu}_{b}(\x, \tf; \Lambda, \alpha) = Q_{\alpha / 2} \left( \left\{ \widehat{\underline{\mu}}(\x, \tf; \Lambda, \params_k)  \right\}_{k = 1}^b \right),
    \end{equation*}
\end{minipage}%
\begin{minipage}{.50\linewidth}
    \begin{equation*}
        \overline{\mu}_{b}(\x, \tf; \Lambda, \alpha) =  Q_{1 - \alpha / 2} \left( \left\{ \widehat{\overline{\mu}}(\x, \tf; \Lambda, \params_k)  \right\}_{k = 1}^b \right).
    \end{equation*}
\end{minipage}

Furthermore, the bootstrap confidence interval of the upper and lower bounds of the APO (dose-response) function under the CMSM are given by:
$
    \label{eq:apo_ci}
    \CI_b\left(\mu(\tf; \Lambda, \alpha) \right) \coloneqq \left[ \underline{\mu}_{b}(\tf; \Lambda, \alpha), \overline{\mu}_{b}(\tf; \Lambda, \alpha) \right], 
$
where,
\noindent
\begin{minipage}{.5\linewidth}
    \begin{equation*}
        \underline{\mu}_{b}(\tf; \Lambda, \alpha) = Q_{\alpha / 2} \left( \left\{ \widehat{\underline{\mu}}(\tf; \Lambda, \params_k)  \right\}_{k = 1}^b \right),
    \end{equation*}
\end{minipage}%
\begin{minipage}{.5\linewidth}
    \begin{equation*}
        \overline{\mu}_{b}(\tf; \Lambda, \alpha) =  Q_{1 - \alpha / 2} \left( \left\{ \widehat{\overline{\mu}}(\tf; \Lambda, \params_k)  \right\}_{k = 1}^b \right).
    \end{equation*}
\end{minipage}

\subsection{Scalable Continuous Treatment Effect Estimation}
\label{sec:scalable}
Following \cite{shalit2017estimating}, \cite{schwab2020learning}, and \cite{nie2021vcnet}, we propose using neural-network architectures with two basic components: a feature extractor, $\bm{\phi}(\x; \params)$ ($\bm{\phi}$, for short) and a conditional outcome prediction block $f(\bm{\phi}, \tf; \params)$. 
The feature extractor design will be problem and data specific.
In \Cref{sec:experiments}, we look at using both a simple feed-forward neural network, and also a transformer \cite{vaswani2017attention}.
For the conditional outcome block, we depart from more complex structures (\cite{schwab2020learning, nie2021vcnet}) and simply focus on a residual \citep{he2016deep}, feed-forward, S-learner \citep{kunzel2019metalearners} structure. 
For the final piece of the puzzle, we follow \cite{jesson2021quantifying} and propose a $n_{\y}$ component Gaussian mixture density:
\begin{equation*}
    p(\y \mid \tf, \x, \params) =
    \sum_{j=1}^{n_{\y}} \pitilde_{j}(\bm{\phi}, \tf; \params) \mathcal{N}\left(\y \mid \mut_{j}(\bm{\phi}, \tf; \params), \sigmat^{2}_j(\bm{\phi}, \tf; \params)\right),
\end{equation*}
and $\mut(\x, \tf; \params) = \sum_{j=1}^{n_{\y}} \pitilde_{j}(\bm{\phi}, \tf; \params)\mut_{j}(\bm{\phi}, \tf; \params)$ \citep{bishop1994mixture}.
Models are optimized by maximizing the log-likelihood of $p(\y \mid \tf, \x, \params)$.

\section{Related Works}
\textbf{Scalable Continuous Treatment Effect Estimation.}
Using neural networks to provide scalable solutions for estimating the effects of continuous-valued interventions has received significant attention in recent years. 
\cite{bica2020estimating} provide a Generative Adversarial Network (GAN) approach. 
The dose-response network (DRNet) \citep{schwab2020learning} provides a more direct adaptation of the TarNet \citep{shalit2017estimating} architecture for continuous treatments.
The varying coefficient network VCNet \citep{nie2021vcnet} generalizes the DRNet approach and provides a formal result for incorporating the target regularization technique presented by \cite{shi2019adapting}.
The RieszNet \citep{chernozhukov2021omitted} provides an alternative approach for targeted regularization.
Adaptation of each method is straightforward for use in our sensitivity analysis framework by replacing the outcome prediction head of the model with a suitable density estimator.

\textbf{Sensitivity and Uncertainty Analyses for Continuous Treatment Effects.}
The prior literature for continuous-valued treatments has focused largely on parametric methods assuming linear treatment/outcome, hidden-confounder/treatment, and hidden-confounder/outcome relationships \citep{carnegie2016assessing, dorie2016flexible, middleton2016bias, oster2019unobservable, cinelli2020making, cinelli2020omitted}. 
In addition to linearity, these parametric methods need to assume the structure and distribution of the unobserved confounding variable(s).
\cite{cinelli2019sensitivity} allows for sensitivity analysis for arbitrary structural causal models under the linearity assumption.
The MSM relaxes both the distributional and linearity assumptions, as does our CMSM extension. 
A two-parameter sensitivity model based on Riesz-Frechet representations of the target functionals, here the APO and CAPO, is proposed by \cite{chernozhukov2021omitted} as a way to incorporate confidence intervals and sensitivity bounds.
In contrast, we use the theoretical background of the marginal sensitivity model to derive a one-parameter sensitivity model.
\cite{detommaso2021causal} purport to quantify the bias induced by unobserved confounding in the effects of continuous-valued interventions, but they do not present a formal sensitivity analysis.
Simultaneously and independently of this work, \cite{marmarelisbounding} are deriving a sensitivity model that bounds the partial derivative of the log density ratio between complete and nominal propensity densities. 
Bounding the effects of continuous valued interventions has also been explored using instrumental variable models \citep{kilbertus2020aclass, hu2021generative, padh2022stochastic}.

\section{Experiments}
\label{sec:experiments}
Here we empirically validate our method.
First, we consider a synthetic structural causal model (SCM) to demonstrate the validity of our method.
Next, we show the scalability of our methods by applying them to a real-world climate-science-inspired problem.
Implementation details (\cref{app:implementation}), datasets (\cref{app:datasets}), and code are provided at 
\url{https://github.com/oatml/overcast}.

\begin{figure}[ht]
    \centering
     \begin{subfigure}[b]{0.22\textwidth}
         \includegraphics[width=\textwidth]{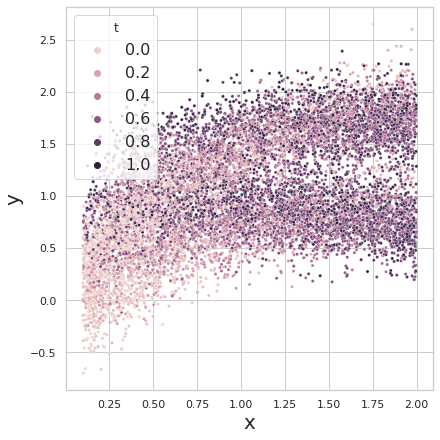}
         \caption{Observed Outcome}
         \label{fig:synthetic_data_outcome}
    \end{subfigure}
    \begin{subfigure}[b]{0.22\textwidth}
        \includegraphics[width=\textwidth]{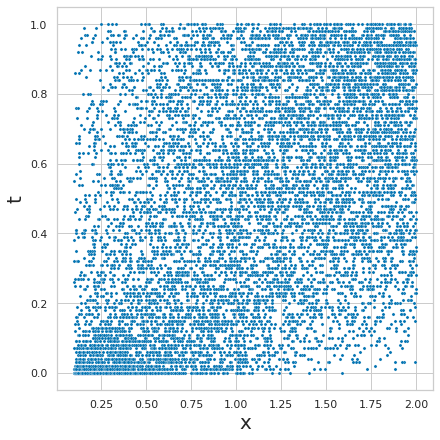}
         \caption{Observed Treatment}
         \label{fig:synthetic_data_treatment}
    \end{subfigure}
    \begin{subfigure}[b]{0.22\textwidth}
        \includegraphics[width=\textwidth]{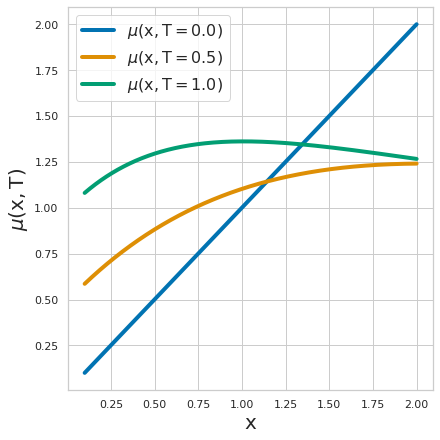}
         \caption{CAPO functions}
         \label{fig:synthetic_data_capo}
    \end{subfigure}
    \begin{subfigure}[b]{0.22\textwidth}
        \includegraphics[width=\textwidth]{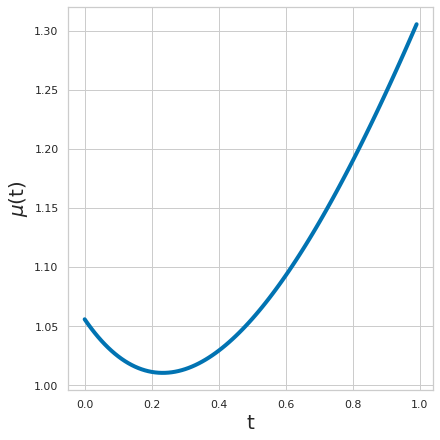}
         \caption{APO function}
         \label{fig:synthetic_data_apo}
    \end{subfigure}
    
    \begin{subfigure}[b]{0.2\textwidth}
        \includegraphics[width=\textwidth]{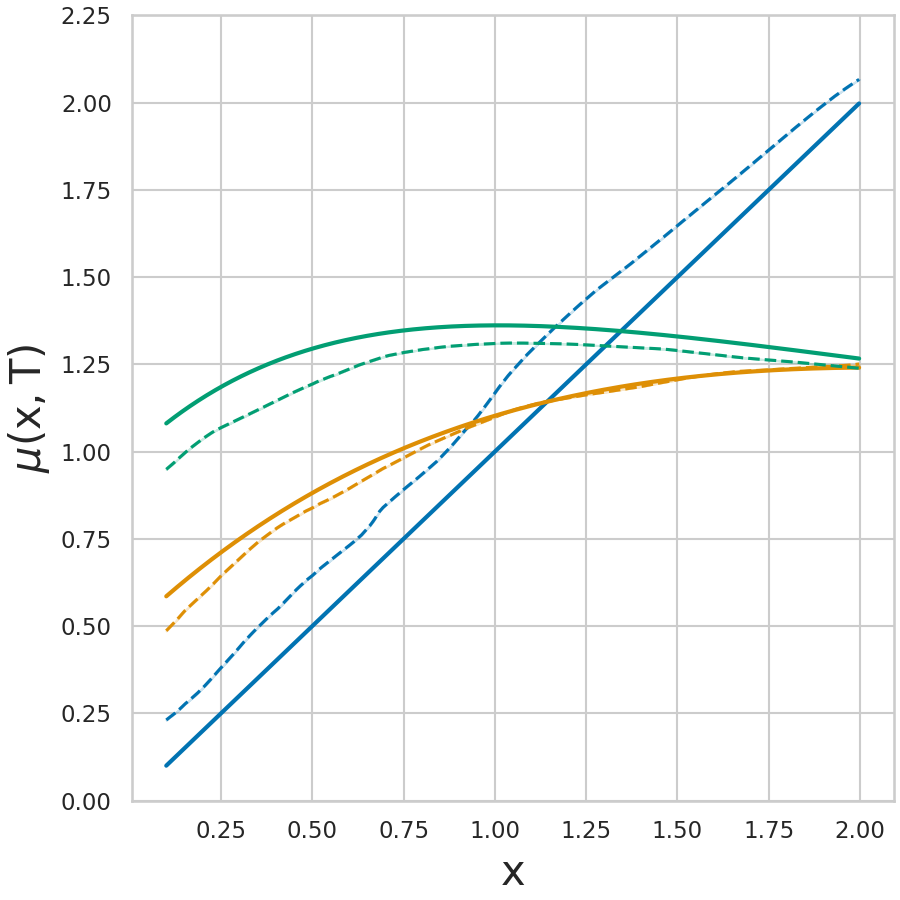}
         \caption{$\Lambda$=1.0}
         \label{fig:capo_causal_1}
    \end{subfigure}
    \begin{subfigure}[b]{0.2\textwidth}
        \includegraphics[width=\textwidth]{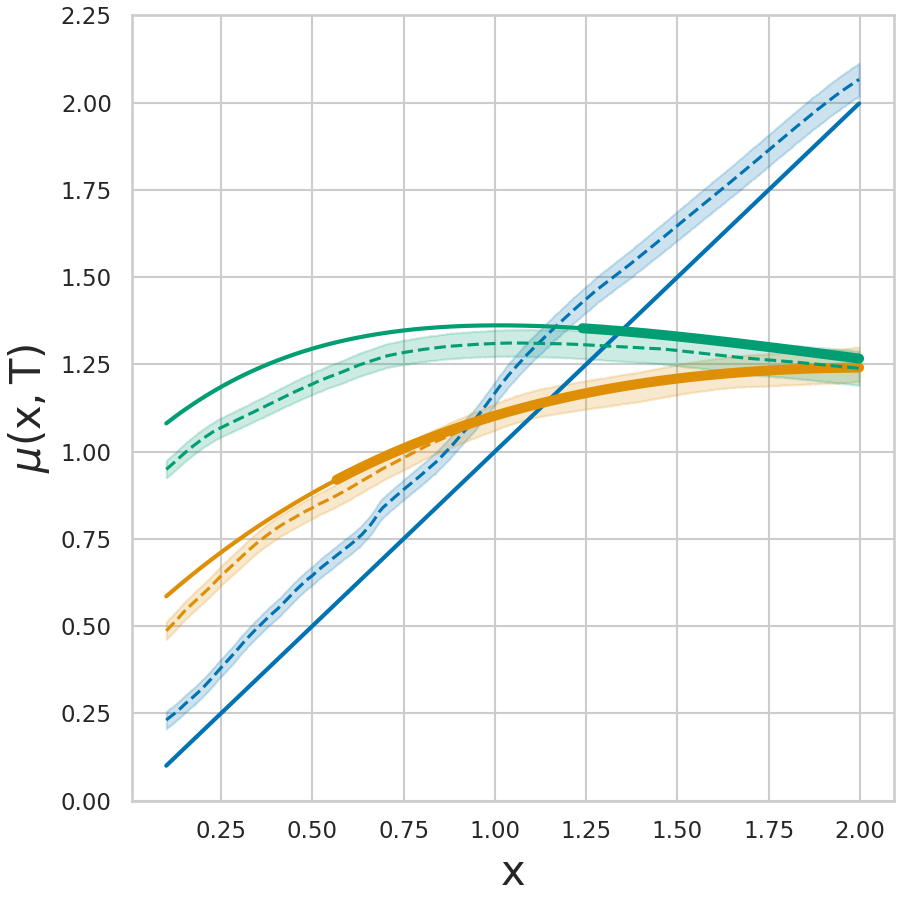}
         \caption{$\Lambda$=1.1}
         \label{fig:capo_causal_2}
    \end{subfigure}
    \begin{subfigure}[b]{0.2\textwidth}
        \includegraphics[width=\textwidth]{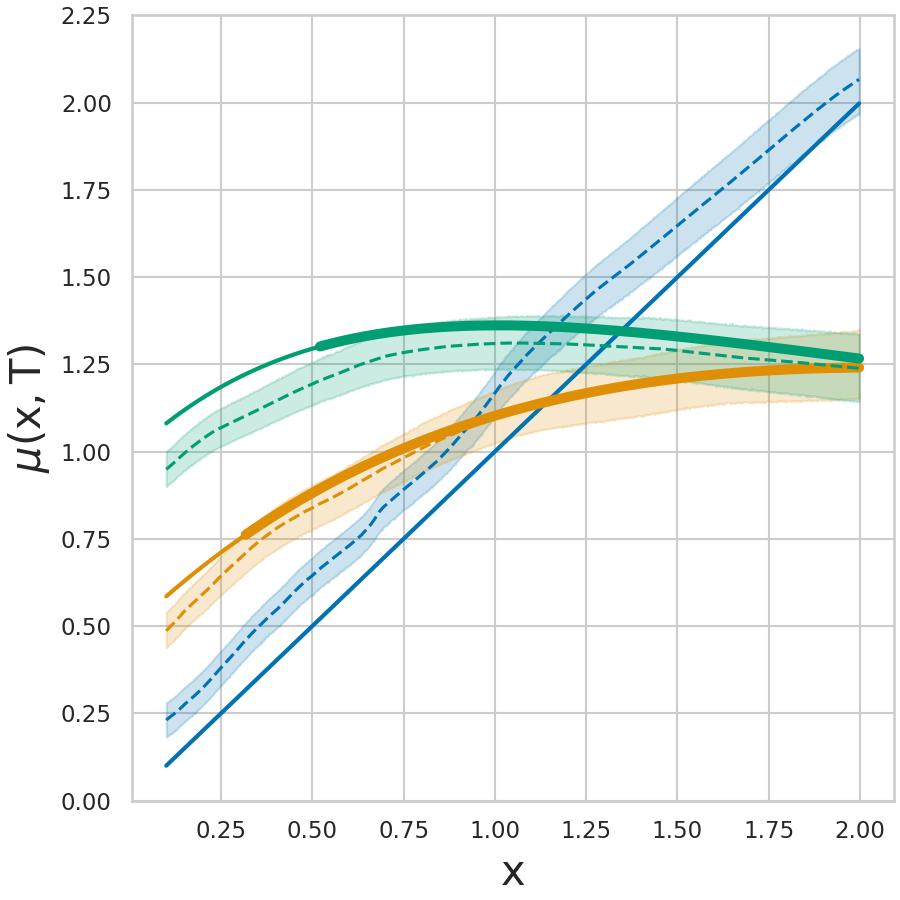}
         \caption{$\Lambda$=1.2}
         \label{fig:capo_causal_3}
    \end{subfigure}
    \begin{subfigure}[b]{0.3\textwidth}
        \includegraphics[width=\textwidth]{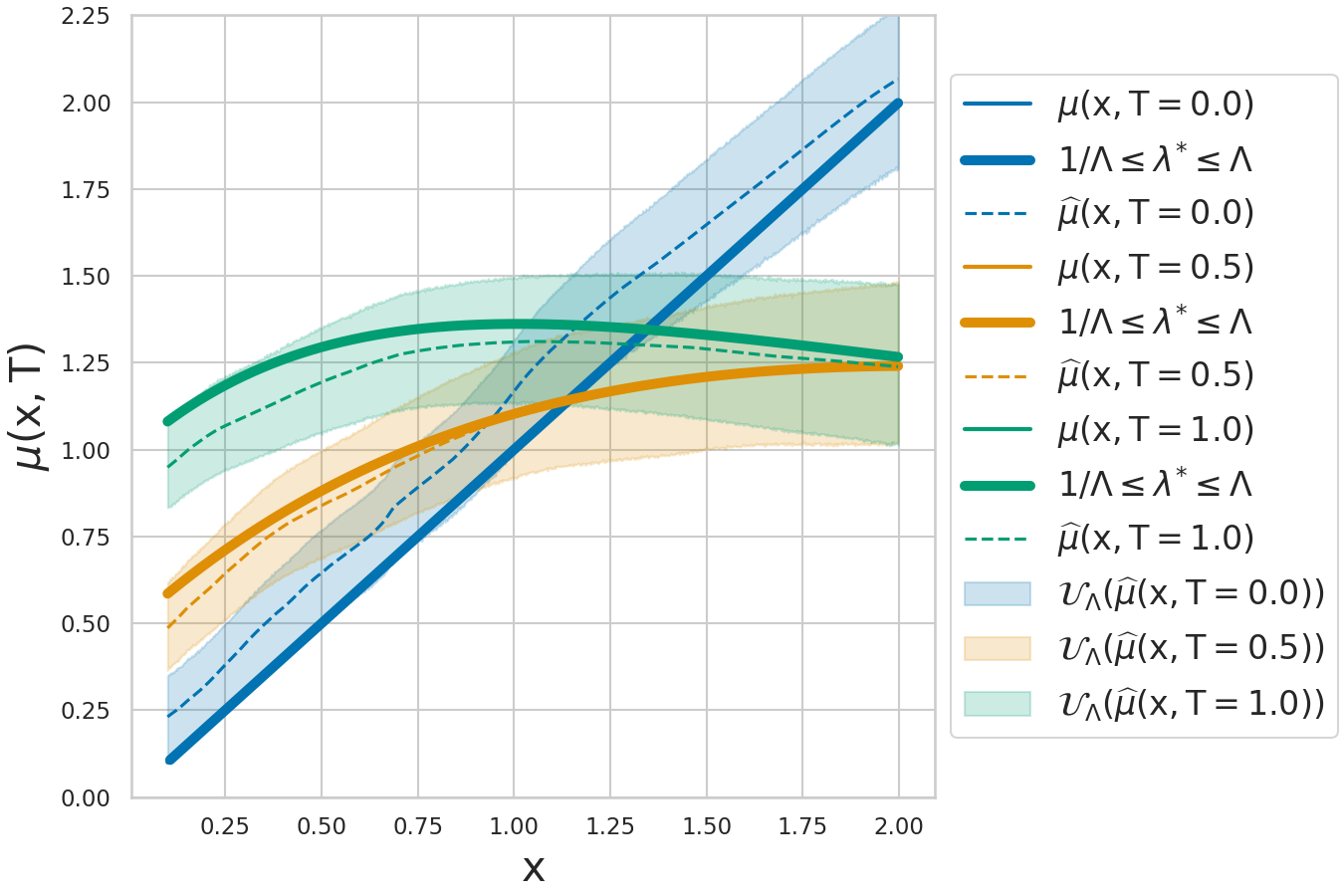}
         \caption{$\Lambda$=1.6}
         \label{fig:capo_causal_4}
    \end{subfigure}
    \caption{
        \Cref{fig:synthetic_data_outcome,fig:synthetic_data_treatment,fig:synthetic_data_capo,fig:synthetic_data_apo}: 
        Synthetic data and ground truth functions.
        \Cref{fig:capo_causal_1,fig:capo_causal_2,fig:capo_causal_3,fig:capo_causal_4}
        Causal uncertainty under hypothesized $\Lambda$ values. 
        Solid lines are ground truth; thick solid lines where the true $\lambda^*$ is within the range of hypothesized $\Lambda$, thin solid lines otherwise. 
        The dotted lines are the estimated CAPO. Shaded regions are estimated CMSM intervals.
    }
    \label{fig:synthetic_data}   
    \vspace{-2em}
\end{figure}
\subsection{Synthetic}
\Cref{fig:synthetic_data} presents the synthetic dataset (additional details about the SCM are given in \Cref{app:synthetic}).
\Cref{fig:synthetic_data_outcome} plots the observed outcomes, $\y$, against the observed confounding covariate, $\mathrm{x}$.
Each datapoint is colored by the magnitude of the observed treatment, $\tf$.
The binary unobserved confounder, $\hu$, induces a bi-modal distribution in the outcome variable, $\y$, at each measured value, $\mathrm{x}$.
\Cref{fig:synthetic_data_treatment} plots the assigned treatment, $\tf$, against the observed confounding covariate, $\mathrm{x}$.
We can see that the coverage of observed treatments, $\tf$, varies for each value of $\mathrm{x}$.
For example, there is uniform coverage at $\mathrm{X}=1$, but low coverage for high treatment values at $\mathrm{X}=0.1$, and low coverage for low treatment values at $\mathrm{X}=2.0$.
\Cref{fig:synthetic_data_capo} plots the true CAPO function over the domain of observed confounding variable, $\mathrm{X}$, for several values of treatment ($\mathrm{T}=0.0$, $\mathrm{T}=0.5$, and $\mathrm{T}=1.0$).
For lower magnitude treatments, $\tf$, the CAPO function becomes more linear, and for higher values, we see more effect heterogeneity and attenuation of the effect size as seen from the slope of the CAPO curve for $\mathrm{T}=0.5$ and $\mathrm{T}=1.0$.
\Cref{fig:synthetic_data_apo} plots the the APO function over the domain of the treatment variable $\mathrm{T}$.

\textbf{Causal Uncertainty}
We want to show that in the limit of large samples (we set $n$ to $100k$), the bounds on the CAPO and APO functions under the CMSM include the ground truth when the CMSM is correctly specified.
That is, when $1 / \Lambda \leq \lambda^*(\tf, \mathrm{x}, \hu) \leq \Lambda$, for user specified parameter $\Lambda$, the estimated intervals should cover the true CAPO or APO.
This is somewhat challenging to demonstrate as the true density ratio $\lambda^*(\tf, \mathrm{x}, \hu)$ (\cref{eq:true_lambda}), varies with $\tf$, $\mathrm{x}$, and $\hu$.
\Cref{fig:capo_causal_1,fig:capo_causal_2,fig:capo_causal_3,fig:capo_causal_4} work towards communicating this.
In \Cref{fig:capo_causal_1}, we see that each predicted CAPO function (dashed lines) is biased away from the true CAPO functions (solid lines). We use thick solid lines to indicate cases where $1 / \Lambda \leq \lambda^*(\tf, \mathrm{x}, \hu) \leq \Lambda$, and thin solid lines otherwise. Therefore thick solid lines indicate areas where we expect the causal intervals to cover the true functions.
Under the erroneous assumption of ignorability ($\Lambda=1$), the CMSM bounds have no width.
In \Cref{fig:capo_causal_2}, we see that as we relax our ignorability assumption ($\Lambda=1.1$) the intervals (shaded regions) start to grow.
Note the thicker orange line: this indicates that for observed data described by $\mathrm{X} > 0.5$ and $\Tf=0.5$, the actual density ratio is in the bounds of the CMSM with parameter $\Lambda=0.5$.
We see that our predicted bounds cover the actual CAPO function for these values.
We see our bounds grow again in \Cref{fig:capo_causal_3} when we increase $\Lambda$ to 1.2.
We see that more data points have $\lambda^*$ values that lie in the CMSM range and that our bounds cover the actual CAPO function for these values. 
In \Cref{fig:capo_causal_4} we again increase the parameter of the CMSM. We see that the bounds grow again, and cover the true CAPO functions for all of the data that satisfy $1 / \Lambda \leq \lambda^*(\tf, \mathrm{x}, \hu) \leq \Lambda$.

\begin{wrapfigure}{r}{0.61\textwidth}
    \vspace{-1em}
    \centering
    \begin{subfigure}[b]{.24\textwidth}
        \includegraphics[width=\textwidth]{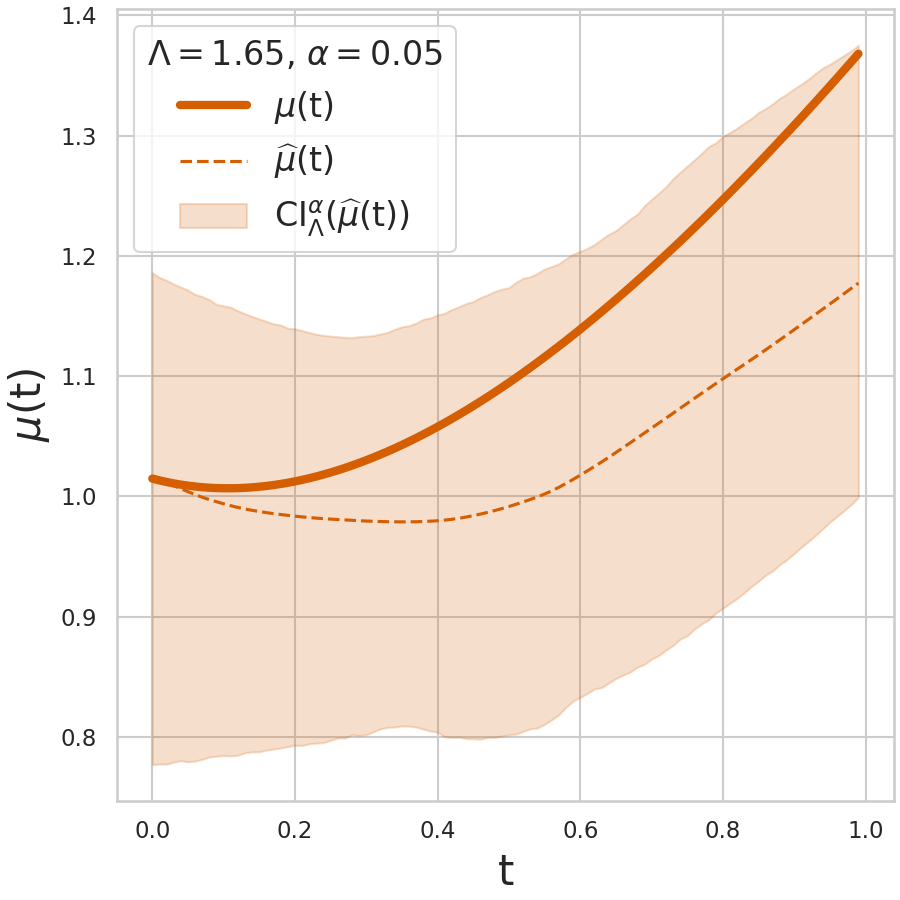}
        \caption{APO Function}
        \label{fig:statcaus_apo}   
    \end{subfigure}
    \begin{subfigure}[b]{0.36\textwidth}  
        \includegraphics[width=\textwidth]{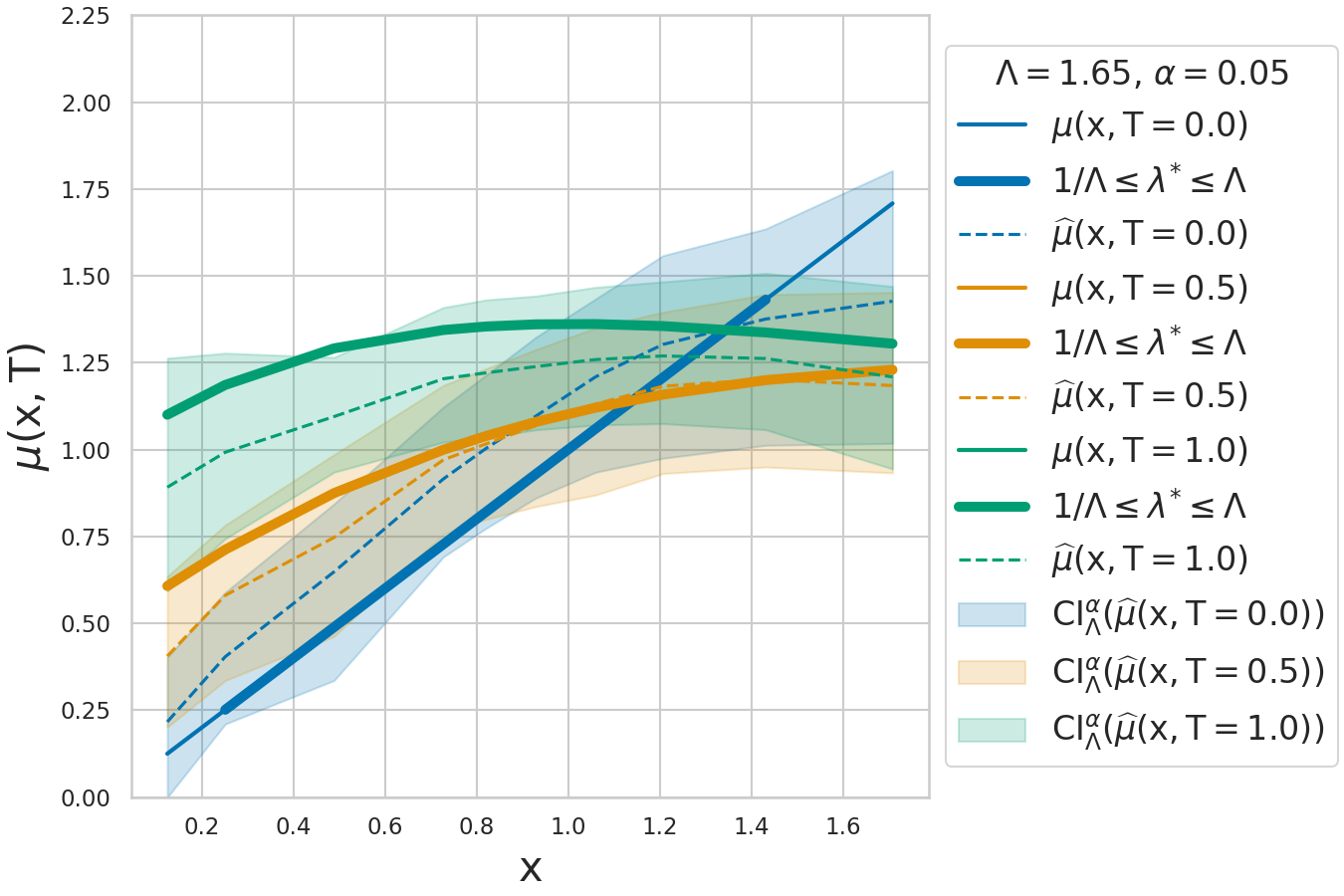}
        \caption{CAPO Functions}
        \label{fig:statcaus_capo} 
    \end{subfigure}
    \caption{Statistical and causal uncertainty, $\alpha$ is statistical significance level for the bootstrap. see  \Cref{fig:synthetic_data} for other details.}
    \label{fig:statcaus}
    \vspace{-1em}
\end{wrapfigure}
\textbf{Statistical Uncertainty}
Now we relax the infinite data assumption and set $n=1000$.
This decrease in data will increase the estimator error for the CAPO and APO functions.
So the estimated functions will not only be biased due to hidden confounding, but they may also be erroneous due to finite sample variance.
We show this in Figure \ref{fig:statcaus_capo} where the blue dashed line deviates from the actual blue solid line as $\mathbf{x}$ increases beyond $1.0$.
However, \Cref{fig:statcaus_capo} shows that under the correct CMSM, the uncertainty aware confidence intervals, \cref{sec:uncertainty}, cover the actual CAPO functions for the range of treatments considered.
\Cref{fig:statcaus_apo} demonstrates that this holds for the APO function as well.

\subsection{Estimating Aerosol-Cloud-Climate Effects from Satellite Data}
\label{sec:clouds}
\textbf{Background}
The development of the model above, and the inclusion of treatment as a continuous variable with multiple, unknown confounders, is motivated by a real-life use case for a prime topic in climate science. Aerosol-cloud interactions (ACI) occur when anthropogenic emissions in the form of aerosol enter a cloud and act as cloud condensation nuclei (CCN). An increase in the number of CCN results in a shift in the cloud droplets to smaller sizes which increases the brightness of a cloud and delays precipitation, increasing the cloud's lifetime, extent, and possibly thickness \citep{twomey1977influence, albrecht1989aerosols, toll2017volcano}. However, the magnitude and sign of these effects are heavily dependent on the environmental conditions surrounding the cloud \citep{douglas2020quantifying}. Clouds remain the largest source of uncertainty in our future climate projections \citep{ipcc6ts}; it is pivotal to understand how human emissions may be altering their ability to cool. Our current climate models fail to accurately emulate ACI, leading to uncertainty bounds that could offset global warming completely or double the effects of rising CO\(_{2}\) \citep{boucher2013clouds}.

\textbf{Defining the Causal Relationships}
Clouds are integral to multiple components of the climate system, as they produce precipitation, reflect incoming sunlight, and can trap outgoing heat \citep{stevens2009untangling}. Unfortunately, their interconnectedness often leads to hidden sources of confounding when trying to address how anthropogenic emissions alter cloud properties.

\begin{figure}[t]
    \centering
    \begin{subfigure}[b]{0.8\textwidth}
        \includegraphics[width=\textwidth]{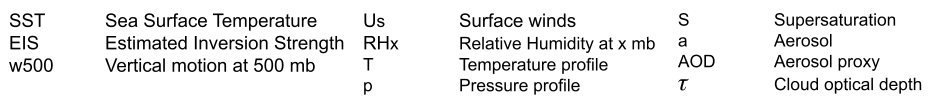}
    \end{subfigure}
    
    \begin{subfigure}[b]{0.4\textwidth}
        \includegraphics[width=\textwidth]{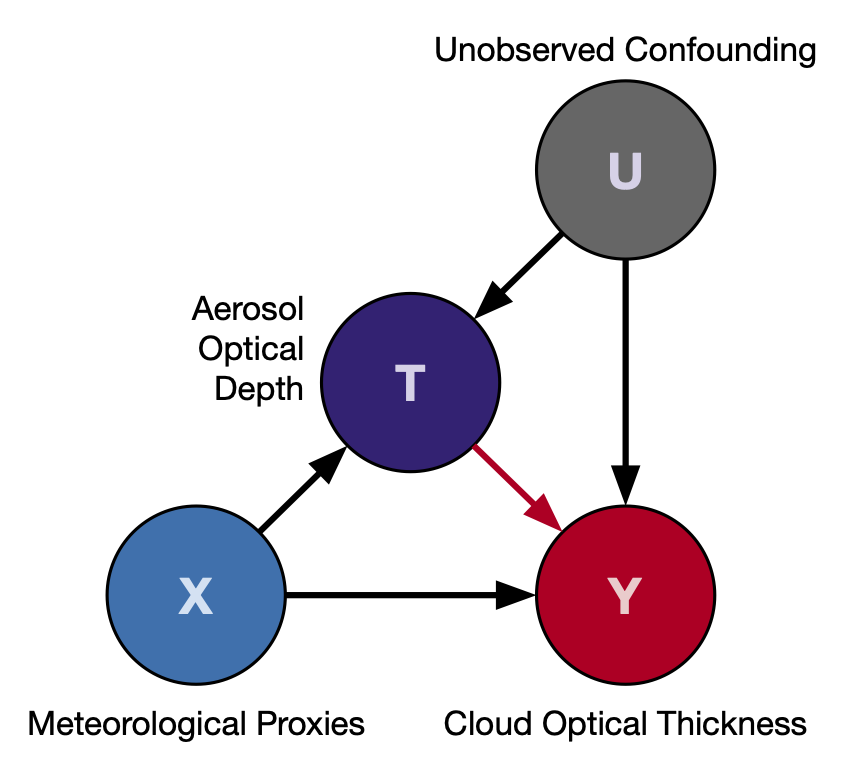}
        \caption{}
        \label{fig:simplecausaldiag}
    \end{subfigure}
    \begin{subfigure}[b]{0.4\textwidth}
        \includegraphics[width=\textwidth]{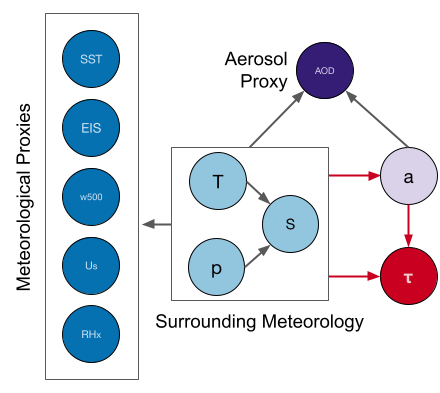}
        \caption{}
        \label{fig:causaldiag}
    \end{subfigure}
    \caption{
        Causal diagrams. \Cref{fig:simplecausaldiag}, a simplified causal diagram representing what we are reporting within; aerosol optical depth (AOD, regarded as the treatment T) modulates cloud optical depth (\(\tau\), Y), which itself is affected by hidden confounders (U) and the meteorological proxies (X).
        \Cref{fig:causaldiag}, an expanded causal diagram of ACI. The aerosol (a) and aerosol proxy (AOD), the true confounders (light blue), their proxies (dark blue), and the cloud optical depth (red).}
    \label{fig:causaldiagram}
    \vspace{-1em}
\end{figure}
Ideally, we would like to understand the effect of aerosols ($a$) on the cloud optical thickness, denoted $\tau$. However, this is currently impossible. Aerosols come in varying concentrations, chemical compositions, and sizes \citep{schutgens2016will} and we cannot measure these variables directly. Therefore, we use aerosol optical depth (AOD) as a continuous, 1-dimensional proxy for aerosols. \Cref{fig:causaldiag} accounts for the known fact that AOD is an imperfect proxy impacted by its surrounding meteorological environment \citep{christensen2017unveiling}.
The meteorological environment is also a confounder that impacts cloud thickness $\tau$ and aerosol concentration $a$. 
Additionally, we depend on simulations of the current environment in the form of reanalysis to serve as its proxy.

Here we report AOD as a continuous treatment and the environmental variables as covariates. 
However, aerosol is the actual treatment, and AOD is only a confounded, imperfect proxy (\Cref{fig:simplecausaldiag}). 
This model cannot accurately capture all causal effects and uncertainty due to known and unknown confounding variables. 
We use this simplified model as a test-bed for the methods developed within this paper and as a demonstration that they can scale to the underlying problem. 
Future work will tackle the more challenging and realistic causal model shown in \Cref{fig:causaldiag}, noting that the treatment of interest $a$ is multi-dimensional and cannot be measured directly.

\begin{figure}[t]
    \centering
     \begin{subfigure}[b]{.4\textwidth}
         \includegraphics[width = \textwidth]{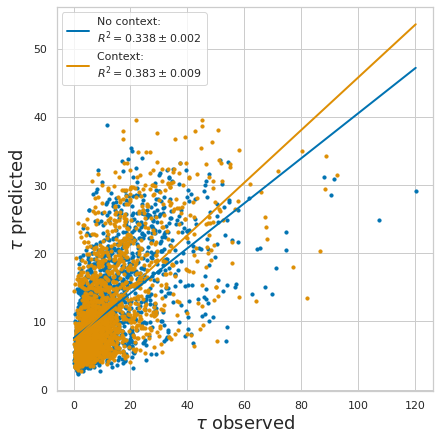}
         \label{fig:CODresults_scatter}
    \end{subfigure}
    \begin{subfigure}[b]{0.4\textwidth}
        \includegraphics[width=\textwidth]{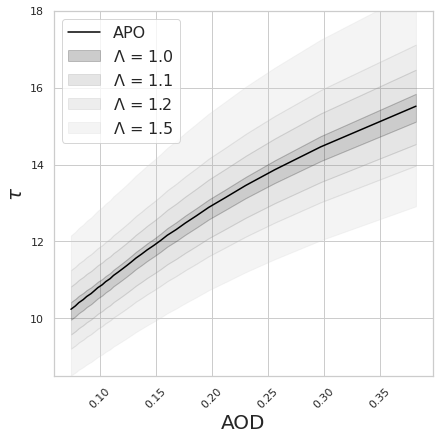}
         \label{fig:CODresults_apo}
    \end{subfigure}
    \caption{Left: The values of the observed, true \(\tau\) against the modeled \(\tau\). Right: The curve for continuous treatment outcome of our aerosol proxy (AOD) on cloud optical depth (\(\tau\)). The darkest shaded region (\(\Lambda\) = 1) represents the uncertainty in the treatment outcome from the ensemble due to finite data. As the strength of confounders increases ($\Lambda > 1.0$), the range of uncertainty in the treatment outcome increases. }
    \label{fig:CODresults}   
    \vspace{-1em}
\end{figure}

\textbf{Model}
We use daily observed $1^{\circ} \times 1^{\circ}$ means of clouds, aerosol, and the environment from sources shown in \Cref{tab:datasources} of \Cref{app:datasets}. 
To model the spatial correlations between the covariates on a given day, we use multi-headed attention \citep{vaswani2017attention} to define a transformer-based feature extractor.
Modeling the spatial dependencies between meteorological variables is motivated by confounding that may be latent in the relationships between neighboring variables.
These dependencies are unobserved from the perspective of a single location.
This architectural change respects both the assumed causal graph (\cref{fig:simplecausaldiag}) and some of the underlying physical causal structure.
We see in \Cref{fig:CODresults} (Left) that modeling \emph{context} with the transformer architecture significantly increases the predictive accuracy of the model when compared to a simple feed-forward neural network (\emph{no context}).
\textbf{Discussion \& Results}
The results for the APO of cloud optical depth (\(\tau\)) as the ``treatment'', AOD, increases are shown in \Cref{fig:CODresults}. 
As the assumed strength of confounding increases (\(\Lambda > 1\)), the range of uncertainty in the treatment outcome increases. 
Within this range of confounding, the modeled outcomes agree with two conflicting hypotheses.
First, that aerosol acts to invigorate the cloud, inducing a large response that would follow a maximum curve within this uncertainty range \citep{christensen2011microphysical, douglas2021global}. 
And second, that aerosol has little impact on cloud depth, and the actual response is a minimal, flat line \citep{gryspeerdt2019constraining}. 
We further find the reported dose-response curves in agreement with multiple estimates of aerosol-cloud interactions using satellite observations \citep{breon2002aerosol, myhre2007aerosol, toll2019weak}. 
The upper bound for $\log \Lambda = .2$ agrees with measurements of the in-cloud environment and aerosol-cloud interactions from aircraft-mounted sensors \citep{painemal2013first}, this may indicate the need for additional control variables when using satellite data.

The resolution of the satellite observations ($1^{\circ} \times 1^{\circ}$ daily means) could be averaging various cloud types and obscuring the signal. 
Future work will investigate how higher resolution (20km $\times$ 20km) data with constraints on cloud type may resolve some confounding influences.
However, even our more detailed causal model (\Cref{fig:causaldiag}) cannot account for all confounders; we expected, and have seen, imperfections in our model of this complex effect. 
The model's results require further expert validation to interpret the outcomes and uncertainty.

\textbf{Societal Impact}
Geoengineering of clouds by aerosol seeding could offset some amount of warming due to climate change, but also have disastrous global impacts on weather patterns \cite{diamond2022opinion}. 
Given the uncertainties involved in understanding aerosol-cloud interactions, it is paramount that policy makers are presented with projected outcomes if a proposals assumptions are wrong or relaxed.

\begin{ack}
We would like to thank Angela Zhou for introducing us to the works of \citep{zhao2019sensitivity} and \citep{dorn2021sharp}. 
These works use the percentile bootstrap for finite sample uncertainty estimation within their sensitivity analysis methods.
We would also like to thank Lewis Smith for helping us understand the Marginal Sensitivity Model of  \cite{tan2006msm} in detail.
Finally, we would like to thank Clare Lyle and all anonymous reviewers for their valuable feedback.

This research was supported by the European Research Council (ERC) project constRaining the EffeCts of Aerosols on Precipitation (RECAP) under the European Union's Horizon 2020 research and innovation program with grant agreement no. 724602 and from the European Union's Horizon 2020 research and innovation program project Constrained aerosol forcing for improved climate projections (FORCeS) under grant agreement No 821205. and Marie Skłodowska-Curie grant agreement No 860100 (iMIRACLI).
This work used JASMIN, the UK's collaborative data analysis environment (http://jasmin.ac.uk).
U.S. was partially supported by the Israel Science Foundation (grant No. 1950/19).
\end{ack}

\bibliography{references}
\bibliographystyle{alpha}

\section*{Checklist}

The checklist follows the references.  Please
read the checklist guidelines carefully for information on how to answer these
questions.  For each question, change the default \answerTODO{} to \answerYes{},
\answerNo{}, or \answerNA{}.  You are strongly encouraged to include a {\bf
justification to your answer}, either by referencing the appropriate section of
your paper or providing a brief inline description.  For example:
\begin{itemize}
  \item Did you include the license to the code and datasets? \answerYes{See Section.}
  \item Did you include the license to the code and datasets? \answerNo{The code and the data are proprietary.}
  \item Did you include the license to the code and datasets? \answerNA{}
\end{itemize}
Please do not modify the questions and only use the provided macros for your
answers.  Note that the Checklist section does not count towards the page
limit.  In your paper, please delete this instructions block and only keep the
Checklist section heading above along with the questions/answers below.

\begin{enumerate}

\item For all authors...
\begin{enumerate}
  \item Do the main claims made in the abstract and introduction accurately reflect the paper's contributions and scope?
    \answerYes{}
    \begin{enumerate}
        \item we claim to introduce a novel marginal sensitivity model for continuous valued treatment effect (the CMSM). See \Cref{sec:methods}.
        \item we claim to derive bounds for the CAPO and APO functions that agree with the CMSMS and observed data. See \Cref{sec:bounds}.
        \item we claim to provide tractable estimators of the CAPO and APO bounds. See \Cref{sec:estimator,sec:algorithm}.
        \item we claim to provide bounds that account for finite-sample (statistical) uncertainty. See \Cref{sec:uncertainty}.
        \item we claim to provide a novel architecture for scalable estimation of the effects of continuous valued interventions. See \Cref{sec:scalable}.
        \item we claim that the bounds cover the true ignorance interval for well specified $\Lambda$. See \Cref{fig:capo_causal_1,fig:capo_causal_2,fig:capo_causal_3,fig:capo_causal_4} and \Cref{th:sharpness}.
        \item we claim that this model scales to real-world, large-sample, high-dimensional data. See \Cref{sec:clouds}
    \end{enumerate}
  \item Did you describe the limitations of your work?
    \answerYes{}
    \begin{enumerate}
        \item We have discussed the major limitation of sensitivity analysis methods, that unobserved confounding is not identifiable from data alone. We have tried to be honest and transparent that our method provides users with a way to communicate the uncertainty induced when relaxing the ignorability assumption.
        We do not claim that lambda is in any way identifiable without further assumptions.
        \item In \Cref{sec:clouds}, we have clearly discussed the limitations of analyses of aerosol-cloud interactions using satellite data where we only see underlying causal mechanisms through proxy variables. We hope this paper serves as a stepping stone for work that specifically addresses those issues.
    \end{enumerate}
  \item Did you discuss any potential negative societal impacts of your work?
    \answerYes{}
  \item Have you read the ethics review guidelines and ensured that your paper conforms to them?
    \answerYes{}
\end{enumerate}

\item If you are including theoretical results...
\begin{enumerate}
    \item Did you state the full set of assumptions of all theoretical results?
        \answerYes{} 
        We have five theoretical results. 
        \Cref{prop:density_ratio}, \Cref{eq:mu_w}, \Cref{eq:CAPO_lower}, \Cref{eq:CAPO_upper}, and \Cref{th:sharpness}.
        All assumptions are stated for each.
    \item Did you include complete proofs of all theoretical results?
        \answerYes{}
        The proof of \Cref{prop:density_ratio} is given in \Cref{proof:density_ratio}. 
        The proof of \Cref{eq:mu_w} is given in \Cref{lem:capo}.
        The proofs for \Cref{eq:CAPO_lower} and \Cref{eq:CAPO_upper} are given in \Cref{lem:monotonic}.
        The proof for \Cref{th:sharpness} is given in \Cref{app:sharpness}.
\end{enumerate}

\item If you ran experiments...
\begin{enumerate}
  \item Did you include the code, data, and instructions needed to reproduce the main experimental results (either in the supplemental material or as a URL)?
    \answerYes{} Code, data, and instructions are provided in the suppleemental material. 
  \item Did you specify all the training details (e.g., data splits, hyperparameters, how they were chosen)?
    \answerYes{} We specify these details in \Cref{app:datasets} and \Cref{app:implementation} as well as in the provided code. 
    \item Did you report error bars (e.g., with respect to the random seed after running experiments multiple times)?
    \answerYes{} Both random seeds and random bootstrapped sampling of the training data.
    \item Did you include the total amount of compute and the type of resources used (e.g., type of GPUs, internal cluster, or cloud provider)?
    \answerYes{} this is outlined in \Cref{app:implementation}
\end{enumerate}

\item If you are using existing assets (e.g., code, data, models) or curating/releasing new assets...
\begin{enumerate}
  \item If your work uses existing assets, did you cite the creators?
    \answerYes{} We use existing satellite data and open source code libraries that we have cited. 
  \item Did you mention the license of the assets?
    \answerYes{}
  \item Did you include any new assets either in the supplemental material or as a URL?
    \answerYes{} we provide a new synthetic dataset and code base
  \item Did you discuss whether and how consent was obtained from people whose data you're using/curating?
    \answerNA{}
  \item Did you discuss whether the data you are using/curating contains personally identifiable information or offensive content?
    \answerNA{}
\end{enumerate}

\item If you used crowdsourcing or conducted research with human subjects...
\begin{enumerate}
  \item Did you include the full text of instructions given to participants and screenshots, if applicable?
    \answerNA{}
  \item Did you describe any potential participant risks, with links to Institutional Review Board (IRB) approvals, if applicable?
    \answerNA{}
  \item Did you include the estimated hourly wage paid to participants and the total amount spent on participant compensation?
    \answerNA{}
\end{enumerate}

\end{enumerate}


\newpage
\appendix
\section{Breaking down the Continuous Treatment Marginal Sensitivity Model}
\label{app:cmsm}
Let's go deeper into the Continuous Treatment Marginal Sensitivity Model (CMSM).

\subsection{MSM for binary treatment values}
\label{app:msm}
This section details the Marginal Sensitivity Model of \cite{tan2006msm}.
For binary treatments, $\Tcal_{B} = \{0, 1\}$, the (nominal) propensity score, $e(\x) \equiv Pr(T=1 \mid \X=\x)$, states how the treatment status, $\tf$, depends on the covariates, $\x$, and is identifiable from observational data.
The potential outcomes, $\Y_{0}$ and $\Y_1$, conditioned on the covariates, $\x$, are distributed as $P(\Y_{0} \mid \X=\x)$ and $P(\Y_{1} \mid \X=\x)$.
Each of these conditional distributions can be written as mixtures with weights based on the propensity score:
\begin{equation}
    \begin{split}
        P(\Y_0 \mid \X=\x) &= (1 - e(\x)) P(\Y_0 \mid \Tf = 0, \X=\x) + e(\x) P(\Y_0 \mid \Tf = 1, \X=\x), \\
        P(\Y_1 \mid \X=\x) &= (1 - e(\x)) P(\Y_0 \mid \Tf = 1, \X=\x) + e(\x) P(\Y_1 \mid \Tf = 1, \X=\x).
    \end{split}
\end{equation}
The conditional distributions of each potential outcome given the observed treatment, $P(\Y_0 \mid \Tf = 0, \X=\x)$ and $P(\Y_1 \mid \Tf = 1, \X=\x)$, are identifiable from observational data, whereas the conditional distributions of each potential outcome given the counterfactual treatment, $P(\Y_0 \mid \Tf = 1, \X=\x)$ and $P(\Y_1 \mid \Tf = 0, \X=\x)$ are not.
Under ignorability, $\{\Y_0, \Y_1\} \indep \Tf \mid \X=\x$, $P(\Y_0 \mid \Tf = 0, \X=\x) = P(\Y_0 \mid \Tf = 1, \X=\x)$ and $P(\Y_1 \mid \Tf = 1, \X=\x) = P(\Y_1 \mid \Tf = 0, \X=\x)$.
Therefore, any deviation from these equalities will be indicative of hidden confounding.
However, because the distributions  $P(\Y_0 \mid \Tf = 1, \X=\x)$ and $P(\Y_1 \mid \Tf = 0, \X=\x)$ are unidentifiable, the MSM postulates a relationship between each pair of identifiable and unidentifiable components.

The MSM assumes that $P(\Y_t \mid \Tf = 1-t, \X=\x)$ is absolutely continuous with respect to $P(\Y_t \mid \Tf = t, \X=\x)$ for all $\tf \in \Tcal_{B}$. Therefore, given that $P(\Y_t \mid \Tf = t, \X=\x)$ and $P(\Y_t \mid \Tf = 1-t, \X=\x)$ are $\sigma$-finite measures, by the Radon-Nikodym theorem, there exists a function $\lambda_{B}(\Yt, \x; \tf): \Ycal \to [0, \inf)$ such that,
\begin{equation}
    P(\Y_t \mid \Tf = 1-t, \X=\x) = \int_{\Ycal} \lambda_{B}(\Yt, \x; \tf) dP(\Y_t \mid \Tf = t, \X=\x).
\end{equation}
Rearranging terms, $\lambda_{B}(\Yt, \x; \tf)$ is expressed as the Radon-Nikodym derivative or ratio of densities,
\begin{equation}
    \begin{split}
        \lambda_{B}(\Yt, \x; \tf) &= \frac{dP(\Y_t \mid \Tf = 1-t, \X=\x)}{dP(\Y_t \mid \Tf = t, \X=\x)}, \\
        &= \frac{p(\y_t \mid \Tf = 1-t, \X=\x)}{p(\y_t \mid \Tf = t, \X=\x)}.
    \end{split}
\end{equation}
By Bayes's rule, $\lambda(\Y_0, \x; 0)$ and $\lambda(\Y_1, \x; 1)$ are expressed as odds ratios,
\begin{equation}
    \begin{split}
        \lambda_{B}(\Y_0, \x; 0) &= \frac{1 - e(\x)}{e(\x)} \bigg/ \frac{1 - e(\x, \y_0)}{e(\x, \y_0)}, \\
        \lambda_{B}(\Y_1, \x; 1) &= \frac{e(\x)}{1 - e(\x)} \bigg/  \frac{e(\x, \y_1)}{1 - e(\x, \y_1)},
    \end{split}
\end{equation}
where $e(\x, \yt) \equiv Pr(\Tf=1 \mid \X=\x, \Yt=\yt)$ is the unidentifiable complete propensity for treatment.

Finally, the MSM further postulates that the odds of receiving the treatment $\Tf=1$ for subjects with covariates $\X=\x$ can only differ from $e(\x) / (1 - e(\x))$ by at most a factor of $\Lambda$,
\begin{equation}
    \Lambda^{-1} \leq \lambda_{B}(\Y_t, \x; t) \leq \Lambda.
\end{equation}

\begin{equation}
    \alpha(e(\x, \tf), \Lambda) = \frac{1}{\Lambda e(\x, \tf)} + 1 - \frac{1}{\Lambda} \leq \frac{1}{e(\x, \tf, \y_t)} \leq \frac{\Lambda}{e(\x, \tf)} + 1 - \Lambda = \beta(e(\x, \tf), \Lambda)
\end{equation}

\subsection{Modifying the MSM for categorical treatment values}
\label{app:cat_msm}
For categorical treatments, $\Tcal_{C} = \{\tf_i\}_{i=1}^{n_c}$, the (nominal) generalized propensity score \cite{hirano2004propensity}, $r(\x, \tf) \equiv Pr(T=\tf \mid \X=\x)$, states how the treatment status, $\tf$, depends on the covariates, $\x$, and is identifiable from observational data.
The potential outcomes, $\{\Y_{\tf}: \tf \in \Tcal_{C}\}$, conditioned on the covariates, $\x$, are distributed as $\{P(\Y_{\tf} \mid \X=\x): \tf \in \Tcal_{C}\}$.
Again, each of these conditional distributions can be written as mixtures with weights based on the propensity density, yielding the following set of mixture distributions:
\begin{equation}
    \left\{
        P(\Y_t \mid \X=\x) = \sum_{\tfp \in \Tcal_{C}} r(\x, \tfp) P(\Yt \mid \Tf=\tfp, \X=\x)
    \right\}.
\end{equation}
Each conditional distribution of the potential outcome given the observed treatment, $P(\Yt \mid \Tf=\tf, \X=\x)$, is identifiable from observational data, but each conditional distribution of the potential outcome given the counterfactual treatment, $P(\Yt \mid \Tf=\tfp, \X=\x)$, and therefore each mixture $P(\Y_t \mid \X=\x)$, is not.
Under the ignorability assumption, $P(\Yt \mid \Tf=\tf, \X=\x) = P(\Yt \mid \Tf=\tfp, \X=\x)$ for all $\tfp \in \Tcal_{C}$.

In order to recover the form of the binary treatment MSM, we can postulate a relationship between the unidentifiable $P(\Y_t \mid \X=\x) - r(\x, \tf) P(\Yt \mid \Tf=\tf, \X=\x)$ and the identifiable $P(\Yt \mid \Tf=\tf, \X=\x) - r(\x, \tf) P(\Yt \mid \Tf=\tf, \X=\x)$. Under the assumption that $P(\Y_t \mid \X=\x) - r(\x, \tf) P(\Yt \mid \Tf=\tf, \X=\x)$ is absolutely continuous with respect to $P(\Yt \mid \Tf=\tf, \X=\x) - r(\x, \tf) P(\Yt \mid \Tf=\tf, \X=\x)$, we define the Radon-Nikodym derivative
\begin{equation}
    \begin{split}
        \lambda_{C}(\Yt, \x; \tf) &= \frac{d(P(\Yt \mid, \X=\x) - r(\x, \tf) P(\Yt \mid \Tf=\tf, \X=\x))}{d (1 - r(\x, \tf)) P(\Y_t \mid \Tf = t, \X=\x)}, \\
        &= \frac{1}{1 - r(\x, \tf)} \left( \frac{dP(\Yt \mid, \X=\x)}{dP(\Y_t \mid \Tf = t, \X=\x)} - \frac{r(\x, \tf)dP(\Y_t \mid \Tf = t, \X=\x)}{dP(\Y_t \mid \Tf = t, \X=\x)} \right), \\
        &= \frac{1}{1 - r(\x, \tf)} \left( \frac{\sum_{\tfp \in \Tcal_{C}} r(\x, \tfp) dP(\Yt \mid \Tf=\tfp, \X=\x)}{dP(\Y_t \mid \Tf = t, \X=\x)} - \frac{r(\x, \tf)dP(\Y_t \mid \Tf = t, \X=\x)}{dP(\Y_t \mid \Tf = t, \X=\x)} \right), \\
        &= \frac{1}{1 - r(\x, \tf)} \left( \frac{\sum_{\tfp \in \Tcal_{C}} r(\x, \tfp) p(\yt \mid \Tf=\tfp, \X=\x)}{p(\y_t \mid \Tf = t, \X=\x)} - \frac{r(\x, \tf)p(\yt \mid \Tf = t, \X=\x)}{p(\yt \mid \Tf = t, \X=\x)} \right), \\
        &= \frac{1}{1 - r(\x, \tf)} \left( \frac{\sum_{\tfp \in \Tcal_{C}} \cancel{r(\x, \tfp)} \frac{p(\Tf=\tfp \mid \yt, \x)\cancel{p(\yt)}}{\cancel{r(\x, \tfp)}}}{\frac{p(\Tf=\tf \mid \yt, \x)\cancel{p(\yt)}}{r(\x, \tf)}} - \frac{r(\x, \tf)\frac{p(\Tf=\tf \mid \yt, \x)\cancel{p(\yt)}}{\cancel{r(\x, \tf)}}}{\frac{p(\Tf=\tf \mid \yt, \x)\cancel{p(\yt)}}{\cancel{r(\x, \tf)}}} \right), \\
        &= \frac{r(\x, \tf)}{1 - r(\x, \tf)} \frac{1 - p(\Tf=\tf \mid \yt, \x)}{p(\Tf=\tf \mid \yt, \x)}, \\
        &= \frac{r(\x, \tf)}{1 - r(\x, \tf)} \bigg{/} \frac{r(\x, \tf, \yt)}{1 - r(\x, \tf, \yt)},
    \end{split}
\end{equation}
where, $r(\x, \tf, \yt) \equiv p(\Tf=\tf \mid \yt, \x)$ is the unidentifiable complete propensity density for treatment.

Finally, the categorical MSM further postulates that the odds of receiving the treatment $\Tf=\tf$ for subjects with covariates $\X=\x$ can only differ from $r(\x, \tf) / (1 - r(\x, \tf))$ by at most a factor of $\Lambda$,
\begin{equation}
    \Lambda^{-1} \leq \lambda_{C}(\Y_t, \x; t) \leq \Lambda.
\end{equation}

\begin{equation}
    \alpha(r(\x, \tf), \Lambda) = \frac{1}{\Lambda r(\x, \tf)} + 1 - \frac{1}{\Lambda} \leq \frac{1}{r(\x, \tf, \y_t)} \leq \frac{\Lambda}{r(\x, \tf)} + 1 - \Lambda = \beta(r(\x, \tf), \Lambda)
\end{equation}


\subsection{Defining the Continuous MSM (CMSM) in terms of densities for continuous-valued interventions}
\label{app:cont_msm}
The conditional distributions of the potential outcomes given the observed treatment assigned, 
\begin{equation*}
    \left\{ P(\Yt \mid \Tf=\tf, \X=\x) : \tf \in \Tcal \right\},
\end{equation*}
are identifiable from observational data. However, the marginal distributions of the potential outcomes over all possible treatments, 
\begin{equation}
    \begin{split}
        \{\quad& \\
            &P(\Yt \mid \X=\x) = \\
            &\quad \int_{\Tcal} p(\tfp \mid \x) P(\Yt \mid \Tf=\tfp, \X=\x) d\tfp  \\
            &: \tf \in \Tcal \\
        \}\quad&
    \end{split}
\end{equation}
are not.
This is because the component distributions, $P(\Yt \mid \Tf=\tfp, \X=\x)$, are not identifiable when $\tfp \neq \tf$ as $\Yt$ cannot be observed for units under treatment level $\Tf = \tfp$.
Under the ignorability assumption, $P(\Yt \mid \Tf=\tf, \X=\x) = P(\Yt \mid \Tf=\tfp, \X=\x)$ for all $\tfp \in \Tcal$, and so $P(\Yt \mid, \X=\x)$ and $P(\Yt \mid \Tf=\tf, \X=\x)$ are identical. 
Therefore, any divergence between $P(\Yt \mid, \X=\x)$ and $P(\Yt \mid \Tf=\tf, \X=\x)$ will be indicative of hidden confounding.

Where in the binary setting the MSM postulates a relationship between the unidentifiable $P(\Yt \mid \Tf=1-\tf, \X=\x)$ and identifiable $P(\Yt \mid \Tf=\tf, \X=\x)$, our CMSM postulates a relationship between the unidentifiable $P(\Yt \mid \X=\x)$ and the identifiable $P(\Yt \mid \Tf=\tf, \X=\x)$. 
\begin{quote}
    The Radon-Nikodym theorem involves a measurable space $(\mathit{X}, \Sigma)$ on which two $\sigma$-finite measures are defined, $\mu$ and $\nu$.'' 
    
    -- Wikipedia
\end{quote}
In our setting, the measurable space is $(\mathbb{R}, \Sigma)$, and our $\sigma$-finite measures are, $\mu=P(\Yt \mid \Tf=\tf, \X=\x)$ and $\nu=P(\Yt \mid \X=\x)$: $\Yt \in \Ycal \subseteq \mathbb{R}$.
\begin{quote}
    If $\nu$ is absolutely continuous with respect to $\mu$ (written $\nu \ll \mu$), then there exists a $\Sigma$-measurable function $f: \mathit{X} \to [0, \infty)$, such that $\nu(\mathit{A}) = \int_{\mathit{A}} f d\mu$ for any measurable set $\mathit{A} \subseteq \textit{X}$. 
    
    -- Wikipedia
\end{quote}
We then need to assume that $P(\Yt \mid \X=\x) \ll P(\Yt \mid \Tf=\tf, \X=\x)$, that is $P(\mathit{A} \mid \Tf=\tf, \X=\x) = 0$ implies $P(\mathit{A} \mid \X=\x) = 0$ for any measurable set $\mathit{A}$. 

This leads us to a proof for \Cref{prop:density_ratio}
\begin{proof}
    \label{proof:density_ratio}
    Further, in our setting we have $f = \lambda(\yt; \x, \tf)$, therefore
    \begin{equation}
        P(\Yt \mid \X=\x) = \int_{\Ycal} \lambda(\yt; \x, \tf) dP(\Yt \mid \Tf=\tf, \X=\x).
    \end{equation}
    
    Let the range of $\Yt$ be the measurable space $(\Ycal, \mathcal{A})$, and $\nu(A)$ denote the Lebesgue measure for any measurable $A \in \mathcal{A}$. 
    Then,
    \begin{subequations}
        \label{eq:dr_proof}
        \begin{align}
            \lambda(\yt; \x, \tf)
            &=\frac{dP(\Yt \mid \X=\x)}{dP(\Yt \mid \Tf=\tf, \X=\x)} \label{eq:dr_proof_a}\\
            &= \frac{dP(\Yt \mid \X=\x)}{d\nu}\frac{d\nu}{dP(\Yt \mid \Tf=\tf, \X=\x)} \label{eq:dr_proof_b} \\
            &= \frac{dP(\Yt \mid \X=\x)}{d\nu}\left(\frac{dP(\Yt \mid \Tf=\tf, \X=\x)}{d\nu}\right)^{-1} \label{eq:dr_proof_c} \\
            &= \frac{d}{d\nu} \int_{A}p(\yt \mid \X=\x)d\nu \left(\frac{d}{d\nu} \int_{A}p(\yt \mid \Tf=\tf, \X=\x)d\nu\right)^{-1} \label{eq:dr_proof_d} \\
            &= \frac{p(\yt \mid \X=\x)}{p(\yt \mid \Tf=\tf, \X=\x)} \label{eq:dr_proof_e} \\
            &= \frac{p(\tf \mid \X=\x)}{p(\tf \mid \Yt=\yt, \X=\x)} \label{eq:dr_proof_f}
        \end{align}
    \end{subequations}
    \Cref{eq:dr_proof_a} by the Radon-Nikodym derivative.
    \Cref{eq:dr_proof_a}-\Cref{eq:dr_proof_c} hold $\nu-$almost everywhere under the assumption $P(\Yt \in A \mid \x) \ll \nu(A) \sim P(\Yt \in A \mid \Tf=\tf, \X=\x)$.
    \Cref{eq:dr_proof_c}-\Cref{eq:dr_proof_d} by the Radon-Nikodym theorem.
    \Cref{eq:dr_proof_d}-\Cref{eq:dr_proof_e} by the fundamental theorem of calculus under the assumption that $p(\yt \mid \x)$ and $p(\yt \mid \Tf=\tf, \X=\x)$ be continuous for $\yt \in \Ycal$.
    \Cref{eq:dr_proof_e}-\Cref{eq:dr_proof_f} by Bayes's Rule.
\end{proof}

The sensitivity analysis parameter $\Lambda$ then bounds the ratio, which leads to our bounds for the inverse complete propensity density:
\begin{equation}
    \begin{split}
        \frac{1}{\Lambda} \leq &\frac{p(\tf \mid \x)}{p(\tf \mid \yt, \x)} \leq \Lambda, \\
        \frac{1}{\Lambda p(\tf \mid \x)} \leq &\frac{1}{p(\tf \mid \yt, \x)} \leq \frac{\Lambda}{p(\tf \mid \x)} \\
        \alpha(p(\tf \mid \x), \Lambda) \leq &\frac{1}{p(\tf \mid \yt, \x)} \leq \beta(p(\tf \mid \x), \Lambda)
    \end{split}
\end{equation}



\subsubsection{KL Divergence}
\label{app:kld}
The bounds on the density ratio can also be expressed as bounds on the Kullback–Leibler divergence between $P(\Yt \mid \Tf=\tf, \X=\x)$ and $P(\Yt \mid \X=\x)$.
\begin{gather}
    \Lambda^{-1} \leq \frac{p(\tf \mid \x)}{p(\tf \mid \yt, \x)} \leq \Lambda, \\
    \log{\left(\Lambda^{-1}\right)}  \leq \log{\left(\frac{p(\tf \mid \x)}{p(\tf \mid \yt, \x)}\right)}  \leq \log{\left(\Lambda\right)} \\
    \E_{p(\y \mid \tf, \x)}\log{\left(\Lambda^{-1}\right)}  \leq \E_{p(\y \mid \tf, \x)}\log{\left(\frac{p(\tf \mid \x)}{p(\tf \mid \yt, \x)}\right)}  \leq \E_{p(\y \mid \tf, \x)}\log{\left(\Lambda\right)} \\
    \log{\left(\Lambda^{-1}\right)}  \leq \E_{p(\y \mid \tf, \x)}\log{\left(\frac{p(\tf \mid \x)}{p(\tf \mid \yt, \x)}\right)}  \leq \log{\left(\Lambda\right)} \\
    \log{\left(\Lambda^{-1}\right)} \leq  \int_{\Ycal}\log\left(\frac{dP(\Yt \mid \X=\x)}{dP(\Yt \mid \Tf=\tf, \X=\x)}\right)dP(\Yt \mid \Tf=\tf, \X=\x)  \leq \log{\left(\Lambda\right)} \\
    \log{\left(\Lambda^{-1}\right)} \leq  -D_{\mathrm{KL}}\left( P(\Yt \mid \Tf=\tf, \X=\x) || P(\Yt \mid \X=\x) \right)  \leq \log{\left(\Lambda\right)} \\
    |\log{\left(\Lambda\right)}| \geq  D_{\mathrm{KL}}\left( P(\Yt \mid \Tf=\tf, \X=\x) || P(\Yt \mid \X=\x) \right)
\end{gather}

\section{Derivation of \Cref{eq:mu_w}}
\begin{lemma}
    \label{lem:capo}
    \begin{align}
        \mu(\x, \tf) = \mut(\x, \tf) + \frac{\int_\Ycal w(\y, \x) (\y - \mut(\x, \tf)) p(\y \mid \tf, \x)d\y}{(\Lambda^2 - 1)^{-1} + \int_\Ycal w(\y, \x)p(\y \mid \tf, \x)d\y}
    \end{align}
    \begin{proof}
        Recall that the conditional average potential outcome, $\mu(\x, \tf) = \E[\Yt \mid \X=\x]$, is unidentifiable without further assumptions. Following \cite{kallus2019interval}, we start from,
        \begin{subequations}
            \begin{align*}
                \mu(\x, \tf) 
                &= \E\left[\Yt \mid \X=\x\right], \\
                &= \frac{\int_{\Ycal}\yt p(\yt \mid \x)d\yt}{\int_{\Ycal}p(\yt \mid \x)d\yt}, \\
                &= \frac{\int_{\Ycal} \yt \frac{p(\tf, \yt \mid \x)}{p(\tf \mid \yt, \x)}d\yt}{\int_{\Ycal} \frac{p(\tf, \yt \mid \x)}{p(\tf \mid \yt, \x)}d\yt}, \\
                &= \frac{\int_{\Ycal} \yt \frac{p(\yt \mid \tf, \x)p(\tf \mid \x)}{p(\tf \mid \yt, \x)}d\yt}{\int_{\Ycal} \frac{p(\yt \mid \tf, \x)p(\tf \mid \x)}{p(\tf \mid \yt, \x)}d\yt}, \\
                &= \frac{\int_{\Ycal} \yt \frac{p(\yt \mid \tf, \x)}{p(\tf \mid \yt, \x)}d\yt}{\int_{\Ycal} \frac{p(\yt \mid \tf, \x)}{p(\tf \mid \yt, \x)}d\yt},
            \end{align*}
        \end{subequations}
        which is convenient as it decomposes $\mu(\x, \tf)$ into it's identifiable, $p(\yt \mid \tf, \x)$, and unidentifiable, $p(\tf \mid \yt, \x)$, parts.
        
        Now, following \cite{jesson2021quantifying}, we add and subtract the empirical conditional outcome $\mut(\x, \tf) = \E[\Y \mid \Tf=\tf, \X=\x]$ from the right-hand-side above:
        \begin{subequations}
            \begin{align}
                \mu(\x, \tf)
                &= \frac{\int_{\Ycal} \yt \frac{p(\yt \mid \tf, \x)}{p(\tf \mid \yt, \x)}d\yt}{\int_{\Ycal} \frac{p(\yt \mid \tf, \x)}{p(\tf \mid \yt, \x)}d\yt}, \\
                &= \mut(\x, \tf) + \frac{\int_{\Ycal} \yt \frac{p(\yt \mid \tf, \x)}{p(\tf \mid \yt, \x)}d\yt}{\int_{\Ycal} \frac{p(\yt \mid \tf, \x)}{p(\tf \mid \yt, \x)}d\yt} - \mut(\x, \tf), \\
                &= \mut(\x, \tf) + \frac{\int_{\Ycal} \yt \frac{p(\yt \mid \tf, \x)}{p(\tf \mid \yt, \x)}d\yt}{\int_{\Ycal} \frac{p(\yt \mid \tf, \x)}{p(\tf \mid \yt, \x)}d\yt} - \mut(\x, \tf) \frac{\int_{\Ycal} \frac{p(\yt \mid \tf, \x)}{p(\tf \mid \yt, \x)}d\yt}{\int_{\Ycal} \frac{p(\yt \mid \tf, \x)}{p(\tf \mid \yt, \x)}d\yt}, \\
                &= \mut(\x, \tf) + \frac{\int_{\Ycal} \yt \frac{p(\yt \mid \tf, \x)}{p(\tf \mid \yt, \x)}d\yt}{\int_{\Ycal} \frac{p(\yt \mid \tf, \x)}{p(\tf \mid \yt, \x)}d\yt} -  \frac{\int_{\Ycal} \mut(\x, \tf) \frac{p(\yt \mid \tf, \x)}{p(\tf \mid \yt, \x)}d\yt}{\int_{\Ycal} \frac{p(\yt \mid \tf, \x)}{p(\tf \mid \yt, \x)}d\yt}, \\
                &= \mut(\x, \tf) + \frac{\int_{\Ycal} (\y - \mut(\x, \tf)) \frac{p(\yt \mid \tf, \x)}{p(\tf \mid \yt, \x)}d\yt}{\int_{\Ycal} \frac{p(\yt \mid \tf, \x)}{p(\tf \mid \yt, \x)}d\yt}.
            \end{align}
        \end{subequations}
        
        Following \cite{kallus2019interval} again, we reparameterize the inverse complete propensity density as, $\frac{1}{p(\tf \mid \yt, \x)} = \alpha(\x; \tf, \Lambda) + w(\y, \x) (\beta(\x; \tf, \Lambda) - \alpha(\x; \tf, \Lambda))$ with $w : \Ycal\times \Xcal \to [0, 1]$. We will shorten this expression to $\frac{1}{p(\tf \mid \yt, \x)} = \alpha + w(\y, \x) (\beta - \alpha)$ below. This gives,
        \begin{subequations}
            \begin{align}
                \mu(\x, \tf)
                &= \mut(\x, \tf) + \frac{\int_{\Ycal} (\y - \mut(\x, \tf)) \frac{p(\yt \mid \tf, \x)}{p(\tf \mid \yt, \x)}d\yt}{\int_{\Ycal} \frac{p(\yt \mid \tf, \x)}{p(\tf \mid \yt, \x)}d\yt}, \\
                &= \mut(\x, \tf) + \frac{\int_{\Ycal} (\alpha + w(\y, \x) (\beta - \alpha))(\y - \mut(\x, \tf)) p(\yt \mid \tf, \x)d\yt}{\int_{\Ycal} (\alpha + w(\y, \x) (\beta - \alpha)) p(\yt \mid \tf, \x)d\yt}, \\
                &= \mut(\x, \tf) + \frac{\alpha\int_{\Ycal} (\y - \mut(\x, \tf)) p(\yt \mid \tf, \x)d\yt + (\beta - \alpha)\int_{\Ycal} (\y - \mut(\x, \tf)) w(\y, \x) p(\yt \mid \tf, \x)d\yt}{\alpha\int_{\Ycal} p(\yt \mid \tf, \x)d\yt + (\beta - \alpha)\int_{\Ycal} w(\y, \x) p(\yt \mid \tf, \x)d\yt}, \\
                &= \mut(\x, \tf) + \frac{\alpha\int_{\Ycal} (\y - \mut(\x, \tf)) p(\yt \mid \tf, \x)d\yt + (\beta - \alpha)\int_{\Ycal} (\y - \mut(\x, \tf)) w(\y, \x) p(\yt \mid \tf, \x)d\yt}{\alpha + (\beta - \alpha)\int_{\Ycal} w(\y, \x) p(\yt \mid \tf, \x)d\yt}, \\
                &= \mut(\x, \tf) + \frac{(\beta - \alpha)\int_{\Ycal} (\y - \mut(\x, \tf)) w(\y, \x) p(\yt \mid \tf, \x)d\yt}{\alpha + (\beta - \alpha)\int_{\Ycal} w(\y, \x) p(\yt \mid \tf, \x)d\yt}, \\
                &= \mut(\x, \tf) + \frac{\int_{\Ycal} (\y - \mut(\x, \tf)) w(\y, \x) p(\yt \mid \tf, \x)d\yt}{\frac{\alpha}{\beta - \alpha} + \int_{\Ycal} w(\y, \x) p(\yt \mid \tf, \x)d\yt}, \\
                &= \mut(\x, \tf) + \frac{\int_{\Ycal} (\y - \mut(\x, \tf)) w(\y, \x) p(\yt \mid \tf, \x)d\yt}{\frac{1 / (\Lambda p(\tf \mid \x))}{\Lambda / p(\tf \mid \x) - 1 / (\Lambda p(\tf \mid \x))} + \int_{\Ycal} w(\y, \x) p(\yt \mid \tf, \x)d\yt}, \\
                &= \mut(\x, \tf) + \frac{\int_{\Ycal} (\y - \mut(\x, \tf)) w(\y, \x) p(\yt \mid \tf, \x)d\yt}{\frac{1}{\Lambda^2 - 1} + \int_{\Ycal} w(\y, \x) p(\yt \mid \tf, \x)d\yt},
            \end{align}
        \end{subequations}
        which concludes the proof.
    \end{proof}
\end{lemma}
\section{Approximating integrals using Gauss-Hermite quadrature}
\label{app:gauss-hermite}
Gauss-Hermite quadrature is a numerical method to approximate indefinite integrals of the following form: $\int_{-\infty}^{\infty}\exp{(-\y^2)f(\y)d\y}$.
In this case,
\begin{equation*}
    \int_{-\infty}^{\infty}\exp{(-\y^2)f(\y)d\y} \approx \sum_{i=1}^{m}g_i f(\y),
\end{equation*}
where $m$ is the number of samples drawn. The $\y_i$ are the roots of the physicists Hermite polynomial $H^*_{m}(\y)(i=1, 2, \dots, m)$ and the weights are given by
\begin{equation*}
    g_i = \frac{2^{m-1}m!\sqrt{\pi}}{m^2[H^*_{m-1}(\y_k)]^2}
\end{equation*}
This method can be used to calculate the expectation of a function, $h(y)$, with respect to a Gaussian distributed outcome $p(\y) = \mathcal{N}(\y \mid \mu, \sigma^2)$ through a change of variables, such that,
\begin{equation}
    \label{eq:gauss-hermite}
    \begin{split}
        \E_{p(\y)}[h(\y)] &= \int_{-\infty}^{\infty} \frac{1}{\sqrt{\pi}}\exp{\left( -\y^2 \right)} h\left( \sqrt{2}\sigma \y + \mu \right) d\y   \\
        &\approx \frac{1}{\sqrt{\pi}} \sum_{i=1}^{m}g_i h\left( \sqrt{2}\sigma \y + \mu \right).
    \end{split}
\end{equation}
\begin{definition}
\label{def:gh-integral-estimator}
{\normalfont Gauss-Hermite quadrature integral estimator when $p(\y | \tf, \x, \params)$ is a parametric Gaussian density estimator, $\mathcal{N}(\y \mid \mut(\x, \tf; \params), \sigmat^{2}(\x, \tf; \params))$}:
    \begin{equation*}
        I_{G}(h(\y)) \coloneqq \frac{1}{\sqrt{\pi}} \sum_{i=1}^{m} g_i h\left( \sqrt{2}\sigmat^{2}(\x, \tf; \params) \y + \mut(\x, \tf; \params) \right)
    \end{equation*}
\end{definition}
Alternatively, when the density of the outcome is modelled using a $n_{\y}$ component Gaussian mixture, $p(\y) = \sum_{j=1}^{n_{\y}} \pi_{j} \mathcal{N}(\y \mid \mu_j, \sigma_{j}^2)$
\begin{equation*}
    \label{eq:gauss-hermite-mix}
    \begin{split}
        \E_{p(\y)}[h(\y)]
        &= \frac{1}{\sqrt{\pi}} \sum_{j=1}^{n_{\y}} \pi_{j} \int_{-\infty}^{\infty} \exp{\left( -\y^2 \right)} h\left( \sqrt{2}\sigma_j \y + \mu_{j} \right) d\y, \\
        &\approx \frac{1}{\sqrt{\pi}} \sum_{j=1}^{n_{\y}} \pi_{j} \sum_{i=1}^{m}g_i h\left( \sqrt{2}\sigma_j \y + \mu_j \right).
    \end{split}
\end{equation*}
\begin{definition}
\label{def:gh-mix-integral-estimator}
{\normalfont Gauss-Hermite quadrature integral estimator for expectations when $p(\y | \tf, \x, \params)$ is a parametric Gaussian Mixture Density, $\sum_{j=1}^{n_{\y}} \pitilde_{j}(\x, \tf; \params) \mathcal{N}\left(\y \mid \mut_{j}(\x, \tf; \params), \sigmat^{2}_j(\x, \tf; \params)\right)$:}
    \begin{equation*}
        I_{GM}(h(\y)) \coloneqq \frac{1}{\sqrt{\pi}} \sum_{j=1}^{n_{\tf}} \pitilde_{j}(\x, \tf; \params) \sum_{i=1}^{m} g_i h\left( \sqrt{2}\sigmat_j(\x, \tf; \params) \y + \mut_{j}(\x, \tf; \params) \right)
    \end{equation*}
\end{definition}
\section{Optimization over step functions}
\label{app:step}
\begin{lemma}
    \label{lem:monotonic}
    The sensitivity bounds given in \Cref{eq:CAPO_lower,eq:CAPO_upper} have the following equivalent expressions:
    \begin{equation*}
        \begin{split}
            \overline{\mu}(\x, \tf; \Lambda) &= \sup_{w(\y) \in \WndH} \mut(\x, \tf) + \frac{\int_\Ycal w(\y) (\y - \mut(\x, \tf)) p(\y \mid \tf, \x)d\y}{(\Lambda^2 - 1)^{-1} + \int_\Ycal w(\y)p(\y \mid \tf, \x)d\y}, \\
            \underline{\mu}(\x, \tf; \Lambda) &= \inf_{w(\y) \in \WniH} \mut(\x, \tf) + \frac{\int_\Ycal w(\y) (\y - \mut(\x, \tf)) p(\y \mid \tf, \x)d\y}{(\Lambda^2 - 1)^{-1} + \int_\Ycal w(\y)p(\y \mid \tf, \x)d\y},
        \end{split}
    \end{equation*}
    where $\mut(\x, \tf) = \E[\Y \mid \X=\x, \Tf=\tf]$, $\WndH = \left\{w: H(\y - \yH) \right\}_{\yH \in \Ycal}$, $\WniH = \left\{w: H(\yH - \y) \right\}_{\yH \in \Ycal}$, and
    \begin{equation*}
        H(\mathrm{z}) \coloneqq 
        \begin{dcases*}
            1, \mathrm{z} \geq 0 \\
            0, \mathrm{z} < 0
        \end{dcases*},
    \end{equation*}
    \begin{proof}
        We follow the argument of \cite{kallus2019interval} and show that our alternative formulations of $\alpha(\cdot, \Lambda)$ and $\beta(\cdot, \Lambda)$ do not change the conclusions of their linear program solution. 
        Starting from $\mu(\x, \tf) = \frac{\int_{\Ycal} \yt \frac{p(\tf, \yt \mid \x)}{p(\tf \mid \yt, \x)}d\yt}{\int_{\Ycal} \frac{p(\tf, \yt \mid \x)}{p(\tf \mid \yt, \x)}d\yt},$ and applying a one-to-one change of variables, $\frac{1}{p(\tf \mid \yt, \x)} = \alpha(\x; \tf, \Lambda) + w(\y) (\beta(\x; \tf, \Lambda) - \alpha(\x; \tf, \Lambda))$ with $w:\Ycal \to [0, 1]$, $\alpha(\x; \tf, \Lambda) = 1 / \Lambda p(\tf \mid \x)$, $\beta(\x; \tf, \Lambda) = \Lambda / p(\tf \mid \x)$, we arrive at:
        \begin{equation}
            \label{eq:upper_capo_app}
            \overline{\mu}(\x, \tf; \Lambda) 
            = \sup_{w: \Ycal \to [0, 1]} \frac{\int_{\Ycal} \y p(\y \mid \tf, \x)d\y + (\lambda^2 - 1)\int_{\Ycal} \y w(\y)p(\y \mid \tf, \x)d\y}{1 + (\lambda^2 - 1)\int_{\Ycal} w(\y)p(\y \mid \tf, \x)d\y},
        \end{equation}
        and
        \begin{equation}
            \label{eq:lower_capo_app}
            \underline{\mu}(\x, \tf; \Lambda) 
            = \inf_{w: \Ycal \to [0, 1]} \frac{\int_{\Ycal} \y p(\y \mid \tf, \x)d\y + (\lambda^2 - 1)\int_{\Ycal} \y w(\y)p(\y \mid \tf, \x)d\y}{1 + (\lambda^2 - 1)\int_{\Ycal} w(\y)p(\y \mid \tf, \x)d\y},
        \end{equation}
        after some cancellations.
        Duality can be used to prove that the $w^*(\y)$ which achieves the supremum in \Cref{eq:upper_capo_app} belongs to the set of step functions $\WndH$. 
        An analogous proof for \Cref{eq:lower_capo_app} would show that the $w^*(\y)$ which achieves the infimum in \Cref{eq:lower_capo_app} belongs to the set of step functions $\WniH$.
        
        The optimization problem in \Cref{eq:upper_capo_app} can be rewritten as a linear-fractional program:
        \begin{subequations}
            \label{eq:lfp}
            \begin{align}
                \text{maximize}\quad & \frac{a\langle \y, w(\y) \rangle_{p(\y \mid \tf, \x)} + c}{b\langle 1, w(\y) \rangle_{p(\y \mid \tf, \x)} + d} \\
                \text{subject to}\quad & 0 \leq w(\y) \leq 1: \forall\y \in \Ycal,
            \end{align}
        \end{subequations}
        where $\langle \cdot, \cdot \rangle_{p(\y \mid \tf, \x)}$ is the inner product with respect to $p(\y \mid \tf, \x)$, $a = b = \lambda^2 - 1$, $c = \int_{\Ycal} \y p(\y \mid \tf, \x) d\y$, and $d = \int_{\Ycal} p(\y \mid \tf, \x) d\y$.
        
        The linear-fractional program of \Cref{eq:lfp} is equivalent to the following linear program:
        \begin{subequations}
            \begin{align}
                \text{maximize}\quad &a\langle \y, \widetilde{w}(\y) \rangle_{p(\y \mid \tf, \x)} + c \widetilde{v}(\x) \\
                \text{subject to}\quad &\widetilde{w}(\y) \leq \widetilde{v}(\x): \forall\y \in \Ycal \label{eq:p_cnstrnt}\\
                &-\widetilde{w}(\y) \leq 0: \forall\y \in \Ycal \label{eq:q_cnstrnt}\\
                & b\langle 1, \widetilde{w}(\y) \rangle_{p(\y \mid \tf, \x)} + d \widetilde{v}(\x) = 1 \label{eq:lambda_cnstrnt}\\
                & \widetilde{v}(\x) \geq 0,
            \end{align}
        \end{subequations}
        where
        \begin{equation*}
            \widetilde{w}(\y) = \frac{w(\y)}{b\langle 1, w(\y) \rangle_{p(\y \mid \tf, \x)} + d} 
            \quad \text{and}
            \quad \widetilde{v}(\x) = \frac{1}{b\langle 1, w(\y) \rangle_{p(\y \mid \tf, \x)} + d}
        \end{equation*}
        by the Charnes-Cooper transformation.
        
        Let the dual function $\rho(\y)$ be associated with the primal constraint \cref{eq:p_cnstrnt}, the dual function $\eta(\y)$ be associated with the primal constraint \cref{eq:q_cnstrnt}, and $\gamma$ be the dual variable associated with the primal constraint \cref{eq:lambda_cnstrnt}. 
        The dual program is then:
        \begin{subequations}
            \begin{align}
                \text{minimize}\quad &\gamma \label{eq:dual_a} \\
                \text{subject to}\quad & \rho(\y) - \eta(\y) + \gamma b p(\y \mid \tf, \x) = a \y p(\y \mid \tf, \x): \forall\y \in \Ycal \label{eq:dual_b} \\
                & -\langle 1, \rho(\y) \rangle + \gamma d \geq c \label{eq:dual_c} \\
                & \rho(\y) \in \mathbb{R}_+, \eta(\y) \in \mathbb{R}_+, \gamma \in \mathbb{R}
            \end{align}
        \end{subequations}
        At most one of $\rho(\y)$ or $\eta(\y)$ is non-zero by complementary slackness; therefore, condition \cref{eq:dual_b} implies that
        \begin{equation*}
            \begin{split}
                \rho(\y) &= (\lambda^2 - 1)p(\y \mid \tf, \x) \max\{\y - \gamma, 0\} \text{ when $\eta=0$,} \\
                \eta(\y) &= (\lambda^2 - 1)p(\y \mid \tf, \x) \max\{\gamma - \y, 0\} \text{ when $\rho=0$.} 
            \end{split}
        \end{equation*}
        \cite{kallus2019interval} argue that constraint \cref{eq:dual_c} ought to be tight (an equivalence) at optimality, otherwise there would exist a smaller, feasible $\gamma$ that satisfies the linear program.
        Therefore,
        \begin{equation}
            \begin{split}
                -\langle 1, \rho(\y) \rangle + \gamma d &= c, \\
                -\int_{\Ycal} (\lambda^2 - 1)p(\y \mid \tf, \x) \max\{\y - \gamma, 0\}d\y + \gamma \int_{\Ycal} p(\y \mid \tf, \x) d\y &=  \int_{\Ycal} \y p(\y \mid \tf, \x) d\y, \\
                (\lambda^2 - 1)\int_{\Ycal} \max\{\y - \gamma, 0\} p(\y \mid \tf, \x)d\y  &= \int_{\Ycal} (\gamma - \y) p(\y \mid \tf, \x) d\y.
            \end{split}
        \end{equation}
        Letting $C_{\Y} > 0$ such that $|\Ycal| \leq C_{\Y}$, it is impossible that either $\gamma > C_{\Y}$ (the r.h.s. would be $0$ and the l.h.s. would be $>0$) or $\gamma < -C_{\Y}$ (the r.h.s. would be $>0$ and the l.h.s. would be $<0$). 
        Thus, $\exists \y^{*} \in [-C_{\Y}, C_{\Y}]$ such that when $\y < \y^{*}$, $\eta > 0$ so $w = 0$ and when $\y \geq \y^{*}$, $\rho > 0$ so $w = 1$. 
        Therefore, the optimal $w^*(\y)$ that achieves the supremum in \Cref{eq:upper_capo_app} is in $\WndH$. 
        
        This result holds under
        \begin{subequations}
            \begin{align}
                \mu(\x, \tf) 
                &= \frac{\int_{\Ycal} \y p(\y \mid \tf, \x)d\y + (\lambda^2 - 1)\int_{\Ycal} \y w(\y)p(\y \mid \tf, \x)d\y}{1 + (\lambda^2 - 1)\int_{\Ycal} w(\y)p(\y \mid \tf, \x)d\y},\\
                &= \frac{\int_{\Ycal} \yt \frac{p(\tf, \yt \mid \x)}{p(\tf \mid \yt, \x)}d\yt}{\int_{\Ycal} \frac{p(\tf, \yt \mid \x)}{p(\tf \mid \yt, \x)}d\yt}, \label{eq:equivalence_b}\\
                &= \mut(\x, \tf) + \frac{\int_\Ycal w(\y) (\y - \mut(\x, \tf)) p(\y \mid \tf, \x)d\y}{(\Lambda^2 - 1)^{-1} + \int_\Ycal w(\y)p(\y \mid \tf, \x)d\y} \label{eq:equivalence_c},
            \end{align}
        \end{subequations}
        thus concluding the proof (\cref{eq:equivalence_b}-\cref{eq:equivalence_c} by \Cref{lem:capo}).
    \end{proof}
\end{lemma}

\subsubsection{Discrete search approaches}
Let $\widehat{\Ycal}=\{\y_i \in \Ycal\}_{i=1}^{k}$ be a set of $k$ values of $\y$, then
\begin{equation*}
    \begin{split}
         \underline{\mu}_{\params}^{H}(\x, \tf) &= \min_{\ystar} \left\{ \widehat{\kappa}_{\params}(\x, \tf; \Lambda, H(\ystar - \y)): \ystar \in \widehat{\Ycal} \right\}, \\
         \overline{\mu}_{\params}^{H}(\x, \tf) &= \max_{\ystar} \left\{ \widehat{\kappa}_{\params}(\x, \tf; \Lambda, H(\y - \ystar)): \ystar \in \widehat{\Ycal} \right\}.
    \end{split}
\end{equation*}

\begin{equation*}
    H(\y) \coloneqq 
    \begin{dcases*}
        1, \y > 0 \\
        0, \y \leq 0
    \end{dcases*}
\end{equation*}

\begin{algorithm}
    \caption{Line Search Interval Optimizer}
    \label{alg:line-search}
    \begin{algorithmic}[1]
        \Require{$\x^{*}$ is an instance of $\X$, $\tf^{*}$ is a treatment level to evaluate, $\Lambda$ is a belief in the amount of hidden confounding, $\params$ are optimized model parameters, $\widehat{\Ycal}$ is a set of unique values $\y \in \Ycal$ \emph{sorted in ascending order}.}
        \Function{LineSearch}{$\x^{*}$, $\tf^{*}$, $\Lambda$, $\params$, $\widehat{\Ycal}$}
            \State $\overline{\mu} \gets -\infty$, $\overline{\kappa} \gets \infty$
            \State $\underline{\mu} \gets \infty$, $\underline{\kappa} \gets -\infty$
            \State $\underline{\delta} \gets $ True, $\overline{\delta} \gets $ True
            \While{$\underline{\delta}$}
                \State $\ystar \gets \textsc{Pop}(\widehat{\Ycal}_c)$ \Comment{$\widehat{\Ycal}_c$ a copy of $\widehat{\Ycal}$}
                \State $\underline{\kappa} \gets \widehat{\kappa}_{\params}(\x, \tf; \Lambda, H(\ystar-\y))$
                \If{$\underline{\kappa} < \underline{\mu}$}
                    \State $\underline{\mu} \gets \underline{\kappa}$
                \Else 
                    \State $\underline{\delta} \gets $ False
                \EndIf
            \EndWhile
            \While{$\overline{\delta}$}
                \State $\ystar \gets \textsc{Pop}(\widehat{\Ycal}_c)$ \Comment{$\widehat{\Ycal}_c$ a copy of $\widehat{\Ycal}$}
                \State $\overline{\kappa} \gets \widehat{\kappa}_{\params}(\x, \tf; \Lambda, H(\y - \ystar))$
                \If{$\overline{\kappa} > \overline{\mu}$}
                    \State $\overline{\mu} \gets \overline{\kappa}$
                \Else 
                    \State $\overline{\delta} \gets $ False
                \EndIf
            \EndWhile
            \State \Return $\underline{\mu}, \overline{\mu}$
        \EndFunction
    \end{algorithmic}
\end{algorithm}
\section{Theorem 1}
\label{app:sharpness}
Assume that 
\begin{enumerate}
    \item $m \to \infty$, \label{ass:m}
    \item $n \to \infty$, \label{ass:n}
    \item $(\X=\x, \Tf=\tf) \in \D_n$, \label{ass:data}
    \item $p(\y \mid \tf, \x, \params)$ converges in measure to $p(\y \mid \tf, \x)$, \label{ass:measure}
    \item $\widetilde{\mu}(\x, \tf; \params)$ is a consistent estimator of $\widetilde{\mu}(\x, \tf)$, \label{ass:prob}
    \item $p(\tf \mid \yt, \x) > 0, \forall \yt \in \Ycal$. \label{ass:pos}
\end{enumerate}
Then, $\underline{\mu}(\x, \tf; \Lambda, \params) \overset{p}{\to} \underline{\mu}(\x, \tf; \Lambda)$ and $\overline{\mu}(\x, \tf; \Lambda, \params) \overset{p}{\to} \overline{\mu}(\x, \tf; \Lambda)$.
\begin{proof}
We prove that $\underline{\mu}(\x, \tf; \Lambda, \params) \overset{p}{\to} \underline{\mu}(\x, \tf; \Lambda)$, from which $\overline{\mu}(\x, \tf; \Lambda, \params) \overset{p}{\to} \overline{\mu}(\x, \tf; \Lambda)$ can be proved analogously. 
Note that $\overset{p}{\to}$ denotes ``convergence in probability''.
We need to show that $\lim_{n}P(| \underline{\mu}(\x, \tf; \Lambda, \params_n) - \underline{\mu}(\x, \tf; \Lambda) | \geq \epsilon)$ = 0, for all $\epsilon > 0$. 
Where $\params_n$ are the model parameters corresponding to a dataset $\D_n$ of $n$ observations.
Recall that,
\[
    \underline{\mu}(\x, \tf; \Lambda) \coloneqq \mut(\x, \tf) + \inf_{w \in \WniH} \frac{\int_\Ycal w(\y) (\y - \mut(\x, \tf)) p(\y \mid \tf, \x)d\y}{(\Lambda^2 - 1)^{-1} + \int_\Ycal w(\y)p(\y \mid \tf, \x)d\y},
\]
and
\[
    \underline{\mu}(\x, \tf; \Lambda, \params_n) \coloneqq \mut(\x, \tf; \params_n) + \inf_{w \in \WniH} \frac{I_m\left( w(\y) (\y - \mut(\x, \tf; \params_n)) \right)}{(\Lambda^2 - 1)^{-1} + I_m\left( w(\y) \right)},
\]
where 
\[
    I_m\left( w(\y) (\y - \mut(\x, \tf; \params_n)) \right) = \frac{1}{m}\sum_{i=1}^{m} w(\y_i) (\y_i - \mut(\x, \tf; \params_n)),
\] 
and 
\[
    I_m\left( w(\y) \right) = \frac{1}{m}\sum_{i=1}^{m} w(\y_i),
\]
with $\y_i \sim p(\y \mid \tf, \x, \params_n)$.

First, by \Cref{ass:m} and the law of large numbers, both
\[
    \lim_{m \to \infty} I_m\left( w(\y) (\y - \mut(\x, \tf; \params_n)) \right) = \int_\Ycal w(\y)(\y - \mut(\x, \tf; \params_n))p(\y \mid \tf, \x; \params_n)d\y,
\]
and
\[
    \lim_{m \to \infty} I_m\left( w(\y) \right) = \int_\Ycal w(\y)p(\y \mid \tf, \x; \params_n)d\y.
\]
Therefore, 
\[
    \lim_{m \to \infty} \underline{\mu}(\x, \tf; \Lambda, \params_n) = \mut(\x, \tf; \params_n) + \inf_{w \in \WniH} \frac{\int_\Ycal w(\y)(\y - \mut(\x, \tf; \params_n))p(\y \mid \tf, \x; \params_n)d\y}{(\Lambda^2 - 1)^{-1} + \int_\Ycal w(\y)p(\y \mid \tf, \x; \params_n)d\y}.
\]
Note that this step was missed by \cite{jesson2021quantifying}.

From here, the proof for Theorem 1 from \cite{jesson2021quantifying} can be followed, substituting in $(\Lambda^2 - 1)^{-1}$ where they write $\alpha^`_{\bm{\omega}}$ and $\alpha^`$.

\end{proof}
\section{Optimization over continuous functions}
\label{app:gradient}

Second, we need a functional estimator for $w(\y, \x)$. We use a neural network, $w(\y, \x; \bm{\omega})$, parameterized by $\bm{\omega}$ with sigmoid non-linearity on the output layer to satisfy the $w : \Ycal\times \Xcal \to [0, 1]$ constraint.

For each $(\Lambda, \tf)$ pair, we then need to solve the following optimization problems:
\begin{equation*}
    \underline{\bm{\omega}} = \argmin_{\bm{\omega}} \frac{1}{n} \sum_{i = 1}^{n} \mu(w(\y, \cdot; \bm{\omega}); \x_i, \tf, \Lambda, \params), \quad \x_i \in \D,
\end{equation*}
and
\begin{equation*}
    \overline{\bm{\omega}} = \argmin_{\bm{\omega}} \frac{1}{n} \sum_{i = 1}^{n} -\mu(w(\y, \cdot; \bm{\omega}); \x_i, \tf, \Lambda, \params), \quad \x_i \in \D,
\end{equation*}
where
\begin{equation*}
    \begin{split}
        \mu&(w(\y, \cdot; \bm{\omega}); \x, \tf, \Lambda, \params) \\
        &\coloneqq \mut(\x, \tf; \params) + \frac{I\left(w(\y, \x; \bm{\omega}) (\y - \mut(\x, \tf; \params))\right)}{(\Lambda^2 - 1)^{-1} + I(w(\y, \x; \bm{\omega}))}.
    \end{split}
\end{equation*}
Each of these problems can then be optimized using stochastic gradient descent \cite{ruder2016overview} and error back-propogation \cite{rumelhart1986learning}.
Since the optimization over $\bm{\omega}$ is non-convex, guarantees on this strategy finding the optimal solution have yet to be established. 
As an alternative, the line-search algorithm presented in \cite{jesson2021quantifying} can also be used with small modifications. 
Under the assumptions of Theorem 1 in \cite{jesson2021quantifying}, with the additional assumption that $\Tf$ is a bounded random variable, we inherit their guarantees on the bound of the conditional average potential outcome.

The upper and lower bounds for the CAPO function under treatment $\Tf=\tf$ and sensitivity parameter $\Lambda$ can be estimated for any observed covariate value, $\X=\x$, as
\begin{equation*}
    \underline{\mu}(\x, \tf; \Lambda, \params) = \mu(w(\y, \cdot; \underline{\bm{\omega}}); \x, \tf, \Lambda, \params),
\end{equation*}
and
\begin{equation*}
    \overline{\mu}(\x, \tf; \Lambda, \params) = \mu(w(\y, \cdot; \overline{\bm{\omega}}); \x, \tf, \Lambda, \params).
\end{equation*}
The upper and lower bounds for the APO (dose-response) function under treatment $\Tf=\tf$ and sensitivity parameter $\Lambda$ can be estimated over any set of observed covariates $\D_\x = \{\x_i\}_{i=1}^n$, as
\begin{equation*}
    \underline{\mu}(\tf; \Lambda, \params) = \frac{1}{n} \sum_{i = 1}^{n} \underline{\mu}(\x_i, \tf; \Lambda, \params), \quad \x_i \in \D_\x,
\end{equation*}
\begin{equation*}
    \overline{\mu}(\tf; \Lambda, \params) = \frac{1}{n} \sum_{i = 1}^{n} \overline{\mu}(\x_i, \tf; \Lambda, \params), \quad \x_i \in \D_\x.
\end{equation*}

\section{Datasets}
\label{app:datasets}
\subsection{Synthetic}
\label{app:synthetic}
\begin{equation}
    \begin{split}
        \hu &\coloneqq N_{\hu}, \\
        \mathrm{x} &\coloneqq N_{\mathrm{x}}, \\
        \tf &\coloneqq N_{\tf}, \\
        \yt &\coloneqq \tf + \x \exp(-\tf \mathrm{x}) - \gamma_{\y}(\hu - 0.5) * (0.5 * \mathrm{x} + 1) + N_{\y},
    \end{split}
\end{equation}
where, 
$N_{\hu} \sim p(\hu) \coloneqq \text{Bern}(\hu \mid 0.5)$, 
$N_{\mathrm{x}} \sim p(\mathrm{x}) \coloneqq \mathrm{Unif}[\mathrm{x} \mid 0.1, 2.0]$, 
$N_{\tf} \sim p(\mathrm{\tf} \mid \mathrm{x}, \mathrm{\hu}) \coloneqq \text{Beta-Binomial}(\tf \mid n=100, \alpha=\mathrm{x} + \gamma_{\tf}\hu, \beta=1)$, 
and $N_{\y} \sim \mathcal{N}(0, 0.04)$.
For the results in this paper $\gamma_t=0.3$ and $\gamma_y=0.5$.

The ground truth ratio, $\lambda = \frac{p(\tf \mid \mathrm{x})}{p(\mathrm{\tf} \mid \mathrm{x}, \mathrm{\hu})}$, is then given by, 
\begin{equation}
\label{eq:true_lambda}
    \begin{split}
        \lambda^*(\tf, \mathrm{x}, \hu) 
        &= \frac{\E_{p(\mathrm{u})}[p(\mathrm{\tf} \mid \mathrm{x}, \mathrm{\hu})]}{p(\mathrm{\tf} \mid \mathrm{x}, \mathrm{\hu})} \\
        &= \frac{\sum_{\hu^{\prime}=0}^1 0.5 * \text{Beta-Binomial}(\tf \mid n=100, \alpha=\mathrm{x} + \gamma_{\tf}\hu^{\prime}, \beta=1) }{\text{Beta-Binomial}(\tf \mid n=100, \alpha=\mathrm{x} + \gamma_{\tf}\hu, \beta=1)}
    \end{split}
\end{equation}

\begin{figure}[ht]
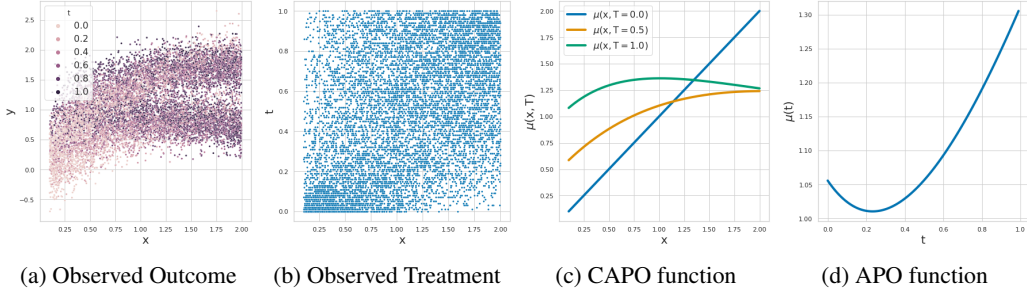

    \centering
     \begin{subfigure}[b]{0.24\textwidth}
         \includegraphics[width=\textwidth]{figures/synthetic/synth_outcome.png}
         \caption{Observed Outcome}
    \end{subfigure}
    \begin{subfigure}[b]{0.24\textwidth}
        \includegraphics[width=\textwidth]{figures/synthetic/synth_treat.png}
         \caption{Observed Treatment}
    \end{subfigure}
    \begin{subfigure}[b]{0.24\textwidth}
        \includegraphics[width=\textwidth]{figures/synthetic/synth_capo.png}
         \caption{CAPO function}
    \end{subfigure}
    \begin{subfigure}[b]{0.24\textwidth}
        \includegraphics[width=\textwidth]{figures/synthetic/synth_apo.png}
         \caption{APO function}
    \end{subfigure}
    \caption{Synthetic data with hidden confounding}
    \label{fig:synthetic_data_app}   
\end{figure}

\subsection{Observations of clouds and aerosol}

\begin{figure}[ht]
    \centering
     \includegraphics[width=.8\textwidth]{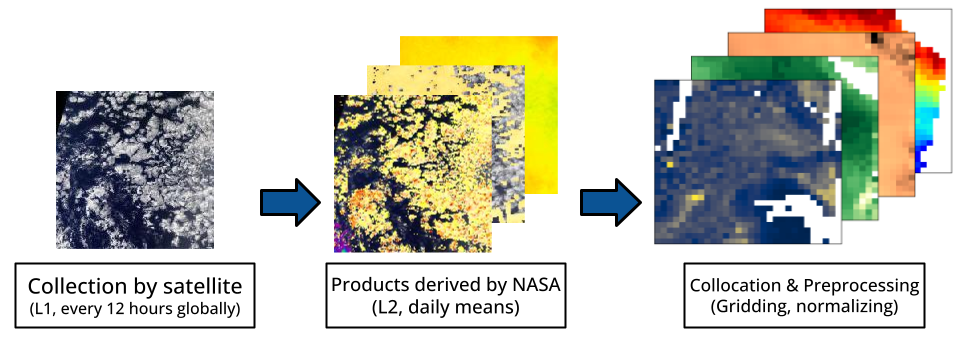}
     \caption{Workflow of observed clouds from satellite to ingestion by model. }
    \label{fig:observed_data_processing}   
\end{figure}

\begin{table}
    \caption{Sources of satellite observations.}
    \label{tab:obstable}
    \centering
    \begin{tabular}{lc}
        \multicolumn{1}{c}{}                   \\
        Product name     & Description \\
        \midrule
        Cloud optical depth \(\tau\) & MODIS   \\
        & (1.6, 2.1, 3.7 \(\mu\)m) \\
        Precipitation & NOAA CMORPH \\
        Sea Surface Temperature & NOAA WHOI \\
        Vertical Motion & MERRA-2 \\
        Estimated Inversion Strength &  MERRA-2  \\
        Relative Humidity & MERRA-2 \\
        Aerosol Optical Depth & MERRA-2  \\
        \bottomrule
        \label{tab:datasources}
    \end{tabular}
    \vspace{-2em}
\end{table}
The Moderate Resolution Imaging Spectroradiometer (MODIS) instrument aboard the Aqua satellite observes the Earth twice daily at \(\sim\)1 km x 1 km resolution native resolution (Level 1) \citep{baum2006introduction}. We used the daily mean, 1\(^{\circ}\) x 1\(^{\circ}\) gridded version (Level 2) in order to somewhat homogenize our observations of clouds and the atmosphere confined to a region off the coast of South America in the Pacific basin. MODIS observations are fed into the Modern-Era Retrospective analysis for Research and Applications version 2 (MERRA-2) real-time model in order to emulate the atmosphere and it's components, such as aerosol \citep{gelaro2017modern}. Aerosol optical depth at 550nm from MERRA-2 is derived from MODIS observations of aerosol from multiple satellites (Terra, Aqua, Suomi-NPP), with corrections for sun glint and near-cloud optical effects \citep{bosilovich2015merra}. We collocated all gridded observations of clouds and reanalysis aerosol with our meteorological proxies of the environment (EIS, SST, w500, RH700, RH850), then normalized our features before feeding them into the model.
\section{Implementation Details}
\label{app:implementation}

Experiments were run using a single NVIDIA GeForce GTX 1080 ti, an Intel(R) Core(TM) i7-8700K, on a desktop computer with 16GB of RAM. 
Code is written in python. Packages used include PyTorch \citep{paszke2019pytorch}, scikit-learn \citep{scikit-learn}, Ray \cite{moritz2018ray}, NumPy, SciPy, and Matplotlib. 
We use ray tune \citep{liaw2018tune} with HyperBand Bayesian Optimization \cite{falkner2017combining} search algorithm to optimize our network hyper-parameters. 
The hyper-parameters we consider are accounted for in Table \ref{tab:search_space}. 
The final hyper-parameters used are given in Table \ref{tab:hparams}.
The hyper-parameter optimization objective is the batch-wise Pearson correlation averaged across all outcomes of the validation data for a single dataset realization with random seed 1331.
All experiments reported can be completed in 30 hours using this setup.

\begin{table}[ht]
    \centering
    \begin{tabular}{ll}
        \toprule
        Hyper-parameter & Search Space \\
        \midrule
        hidden units        & tune.qlograndint(32, 512, 32) \\
        network depth       & tune.randint(2, 5) \\
        gmm components      & tune.randint(1, 32) \\
        attention heads     & tune.randint(1, 8) \\
        negative slope      & tune.quniform(0.0, 0.5, 0.01) \\
        dropout rate        & tune.quniform(0.0, 0.5, 0.01) \\
        layer norm          & tune.choice([True, False]) \\
        batch size          & tune.qlograndint(32, 256, 32) \\
        learning rate       & tune.quniform(1e-4, 1e-3, 1e-4) \\
        \bottomrule
    \end{tabular}
    \caption{Hyper-parameter search space}
    \label{tab:search_space}
\end{table}

\begin{table}[ht]
    \centering
    \begin{tabular}{lccc}
        \toprule
        Hyper-parameter & Synthetic & ACCE NN & ACCE Transformer \\
        \midrule
        hidden units        & 96   & 256   & 256 \\
        network depth       & 4     & 3     & 3 \\
        gmm components      & 24    & 24    & 24 \\
        attention heads     & NA    & NA    & 4 \\
        negative slope      & 0.05  & 0.04  & 0.01 \\
        dropout rate        & 0.04  & 0.2  & 0.5 \\
        layer norm          & False  & False  & False \\
        batch size          & 32    & 2048   & 32 \\
        learning rate       & 0.0015  & 1e-4  & 2e-4 \\
        \bottomrule
    \end{tabular}
    \caption{Final hyper-parameters for each dataset/model}
    \label{tab:hparams}
\end{table}

\subsection{Model Architecture}

\begin{figure}[ht]
    \centering
    \includegraphics[width=\textwidth]{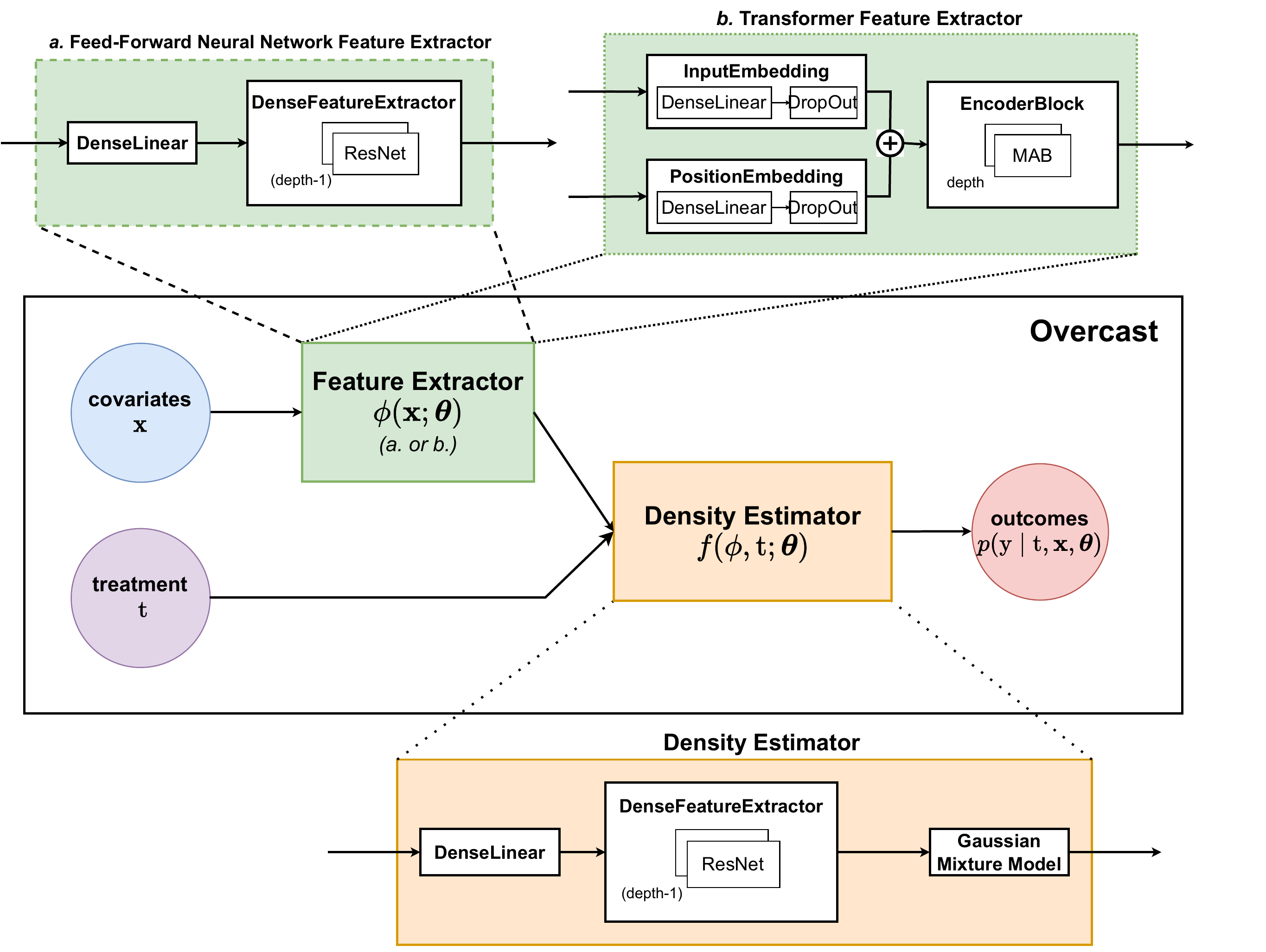}
    \caption{\textbf{Overcast model architecture.} The inputs are represented by circles, in \textcolor{blue}{\textbf{blue}} the covariates, and in \textcolor{purple}{\textbf{purple}} the treatment. In the \textcolor{red}{\textbf{red}} circle is the output of the model, the outcomes distribution. The model has different feature extractors (in \textcolor{green}{green}) for the feed-forward neural network and the transformer. It has a single density estimator (in \textcolor{orange}{orange}).}
    \label{fig:overcast-architecture}
\end{figure}

The general model architecture is shown in \Cref{fig:overcast-architecture}. 
The models are neural-network architectures with two basic components: a feature extractor, $\bm{\phi}(\x; \params)$ ($\bm{\phi}$, for short), and a conditional outcome prediction block $f(\bm{\phi}, \tf; \params)$, or density estimator. 
The covariates $\x$ (represented in \textcolor{blue}{\textbf{blue}}) are given as input to the feature extractor, whose output is concatenated with the treatment $\tf$ (represented in \textcolor{purple}{\textbf{purple}}) and given as input to the density estimator which outputs a Gaussian mixture density $p(\y \mid \tf, \x, \params)$ from which we can sample to obtain samples of the outcomes (represented in \textcolor{red}{\textbf{red}}).
Models are optimized by maximizing the log-likelihood, $\log{p(\y \mid \tf, \x, \params)}$, using mini-batch stochastic gradient descent.

\subsubsection{Feature extractor}

The feature extractor design is problem and data specific. 
In our case, we look at using both a simple feed-forward neural network and also a transformer. 
The transformer has the advantage of allowing us to model the spatio-temporal correlations between the covariates on a given day using the geographical coordinates of the observations as positional encoding.
This is interesting when studying ACI because confounding may be latent in the relationships between neighboring variables.
Typically, environmental processes (which is one source of confounding) are dependent upon the spatial distribution of clouds, humidity and aerosol, and this feature extractor may capture these confounding effects better. 

\subsection{Density Estimator}

\begin{figure}
    \centering
    \includegraphics[width=.6\linewidth]{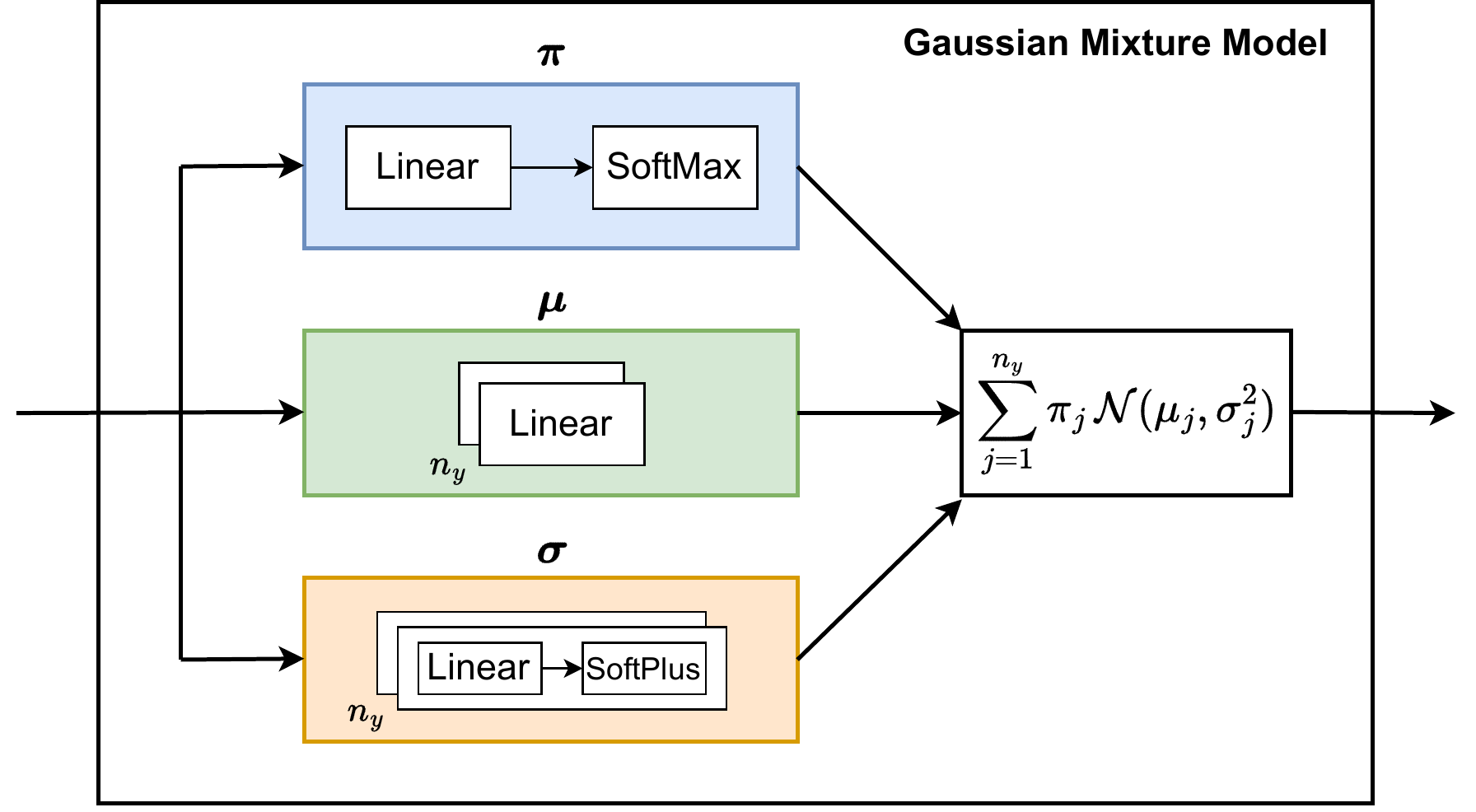}
    \caption{\textbf{Overcast Gaussian mixture model.} The mixing coefficients $\mathbf{\pi}$ are estimated with a linear layer and a SoftMax layer, to obtain $\tilde{\mathbf{\pi}}$, represented in \textcolor{blue}{\textbf{blue}} in the figure. The vector of means of the Gaussian kernels $\tilde{\mathbf{\mu}}$ is obtained by $n_y$ linear layers (in \textcolor{green}{\textbf{green}} in the diagram), whilst the vector of variances $\tilde{\mathbf{\sigma}}$ is obtained by $n_y$ blocks of linear layers and SoftPlus layers (in \textcolor{orange}{\textbf{orange}} in the diagram).}
    \label{fig:overcast-gmm}
\end{figure}

The conditional outcome prediction block, relies on a $n_{\y}$ component Gaussian mixture density represented in \Cref{fig:overcast-gmm}.
It outputs: 
\begin{equation*}
    p(\y \mid \tf, \x, \params) =
    \sum_{j=1}^{n_{\y}} \pitilde_{j}(\bm{\phi}, \tf; \params) \mathcal{N}\left(\y \mid \mut_{j}(\bm{\phi}, \tf; \params), \sigmat^{2}_j(\bm{\phi}, \tf; \params)\right),
\end{equation*}
and 
$$
\mut(\x, \tf; \params) = \sum_{j=1}^{n_{\y}} \pitilde_{j}(\bm{\phi}, \tf; \params)\mut_{j}(\bm{\phi}, \tf; \params),
$$
where $\mathcal{N}(\cdot \mid \mu, \sigma^2)$ denotes a normal distribution with mean $\mu$ and variance $\sigma^2$. 

\section{Additional Results}
\subsection{Synthetic}
\begin{figure}[ht]
    \centering
     \begin{subfigure}[b]{.32\textwidth}
        \includegraphics[width = \textwidth]{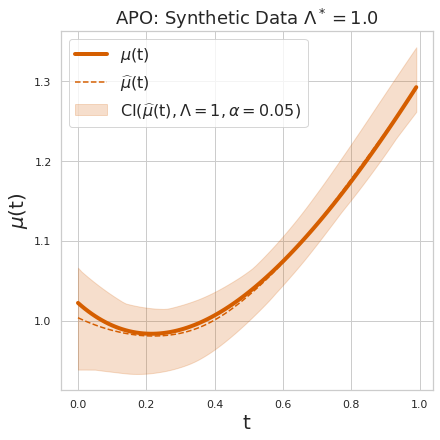}
        \caption{APO Functions}
        \label{fig:statistical-uncertainty-apo}   
    \end{subfigure}
    \begin{subfigure}[b]{0.54\textwidth} 
        \includegraphics[width=\textwidth]{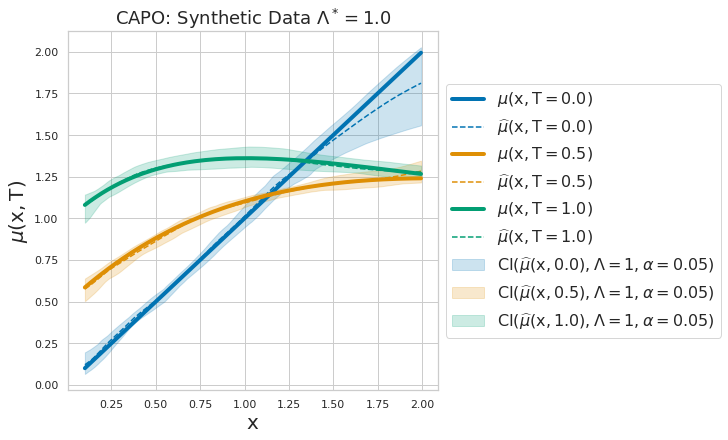}
        \caption{CAPO Functions}
        \label{fig:statistical-uncertainty-capo}  
    \end{subfigure}
    \caption{Investigating statistical uncertainty using unconfounded synthetic data.}
    \label{fig:statistical-uncertainty}   
\end{figure}

\subsection{Aerosol-Cloud-Climate Effects}

\begin{figure}
    \centering
    \includegraphics[width=\textwidth]{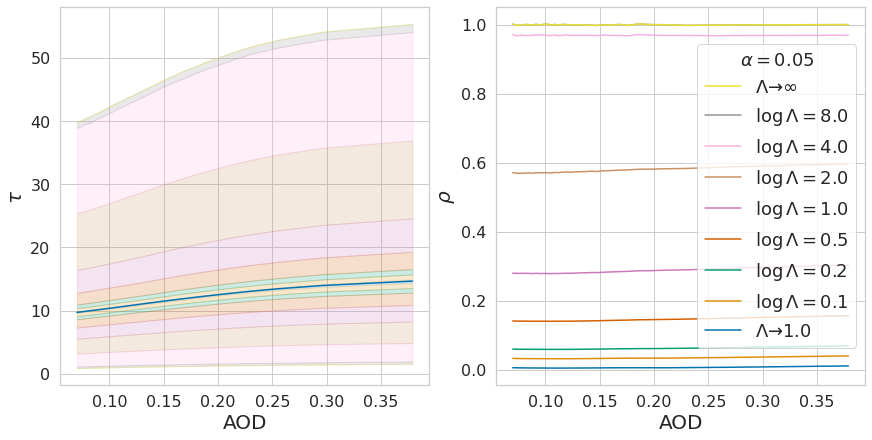}
    \caption{Interpreting $\Lambda$ as a proportion ($\rho$) of the unexplained range of $\Yt$ attributed to unobserved confounding variables.}
    \label{fig:rho}
\end{figure}

In \Cref{fig:rho} we show how $\Lambda$ can be interpreted as the proportion, $\rho$, of the unexplained range of $\Yt$ attributed to unobserved confounding variables. 
In the left figure, we plot the corresponding bounds for increasing values of $\Lambda$ of the predicted AOD-$\tau$ dose-response curves. 
In the right figure we plot the $\rho$ value for each $\Lambda$ at each value of $\tf$. For the curves reported in \Cref{fig:CODresults_apo}: we find that $\Lambda = 1.1$ leads to $\rho \approx 0.04$, $\Lambda = 1.2$ leads to $\rho \approx 0.07$, and $\Lambda = 1.6$ leads to $\rho \approx 0.15$. 
This shows that when we let just a small amount of the unexplained range of $\Yt$ be attributed to unobserved confounding, the  range  of the predicted APO curves become quite wide. 
If we were to completely relax the no-hidden-confounding assumption, the entire range seen in \Cref{fig:rho} 
Left would be plausible for the APO function. 
This range dwarfs the predicted APO curve. 
These results highlight the importance of reporting such sensitivity analyses.

\begin{figure}[ht]
    \centering
     \begin{subfigure}[b]{.32\textwidth} 
        \includegraphics[width = \textwidth]{figures/COD/dr_tau_attn.png}
        \caption{}
    \end{subfigure}
    \begin{subfigure}[b]{0.32\textwidth}
        \includegraphics[width = \textwidth]{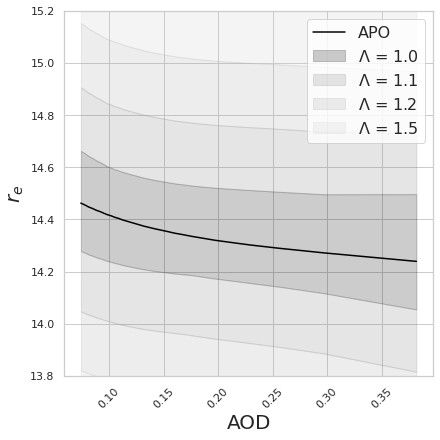}
        \caption{}
    \end{subfigure}
    \begin{subfigure}[b]{0.32\textwidth}
        \includegraphics[width = \textwidth]{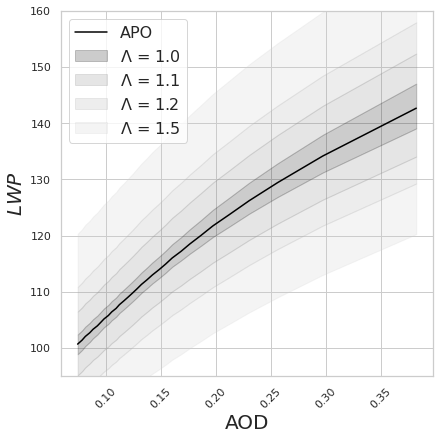}
        \caption{}
    \end{subfigure}
    \caption{Average dose-response curves for other cloud properties. a) Cloud optical depth. b) Water droplet effective radius. c) Liquid water path.}
    \label{fig:cloud-property-dr-curves}
\end{figure}

In \Cref{fig:cloud-property-dr-curves} we show additional dose response curves for cloud optical thickness ($\tau$), water droplet effective radius ($r_e$), and liquid water path (LWP). 

\begin{figure}[ht]
    \centering
     \begin{subfigure}[b]{.32\textwidth} 
        \includegraphics[width = \textwidth]{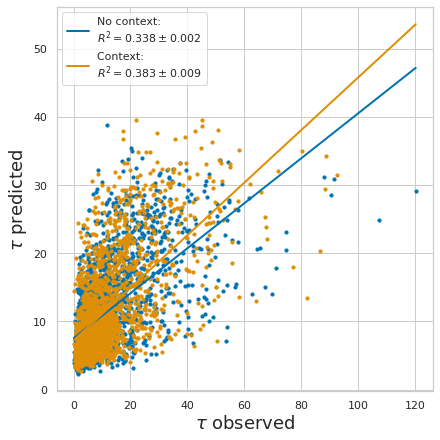}
        \caption{}
    \end{subfigure}
    \begin{subfigure}[b]{0.32\textwidth}
        \includegraphics[width = \textwidth]{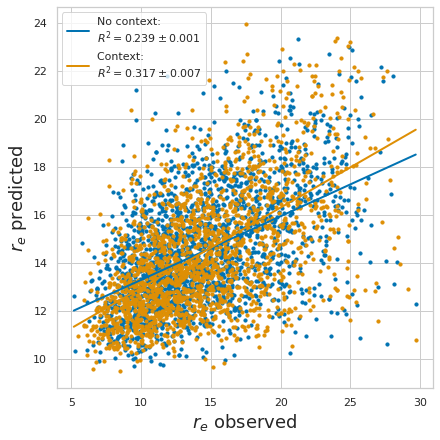}
        \caption{}
    \end{subfigure}
    \begin{subfigure}[b]{0.32\textwidth}
        \includegraphics[width = \textwidth]{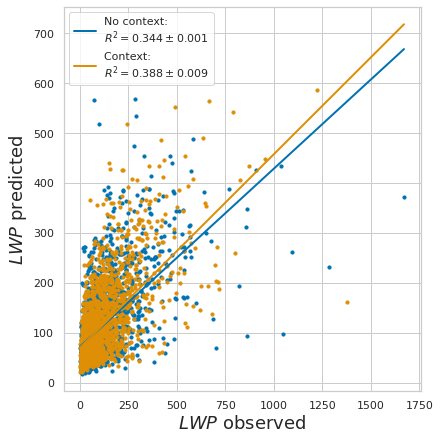}
        \caption{}
    \end{subfigure}
    \caption{Comparing transformer to feed-forward feature extractor at predicting cloud properties given covariates and AOD. a) Cloud optical depth. b) Water droplet effective radius. c) Liquid water path. We see a significant improvement in pearson correlation ($R^2$) in each case.}
    \label{fig:cloud-property-scatterplots}
\end{figure}

In \Cref{fig:cloud-property-scatterplots} we show additional scatter plots comparing the neural network and transformer models for cloud optical thickness ($\tau$), water droplet effective radius ($r_e$), and liquid water path (LWP). 

\subsection{$\omega_{500}$ experiment}

The Overcast models make use of expert knowledge about ACI to select the covariates. 
Ideally, they would include pressure profiles, temperature profiles and supersaturation since these are directly involved in cloud processes and impact the quality of AOD measurements as a proxy for aerosol concentration. 
Unfortunately, they are impossible to retrieve from satellite data, so we rely on meteorological proxies like relative humidity, sea surface temperature, inversion strengths, and vertical motion. 
Relying on these proxies however results in ignorability violations, which must be accounted for in the parameter $\Lambda$ in order to derive appropriate plausible ranges of outcomes.

In the experiment that follows, we are removing a confounding variable from the model, therefore inducing hidden confounding.
The covariate we remove is vertical motion at 500 mb, denoted by $\omega500$. 
This experiment helps us gain some intuition about the influence of the parameter $\Lambda$ and how it relates to the inclusion of confounding variables in the model. 

\begin{figure}[htbp]
    \centering
    \begin{subfigure}[b]{0.46\textwidth} 
        \includegraphics[width = \textwidth]{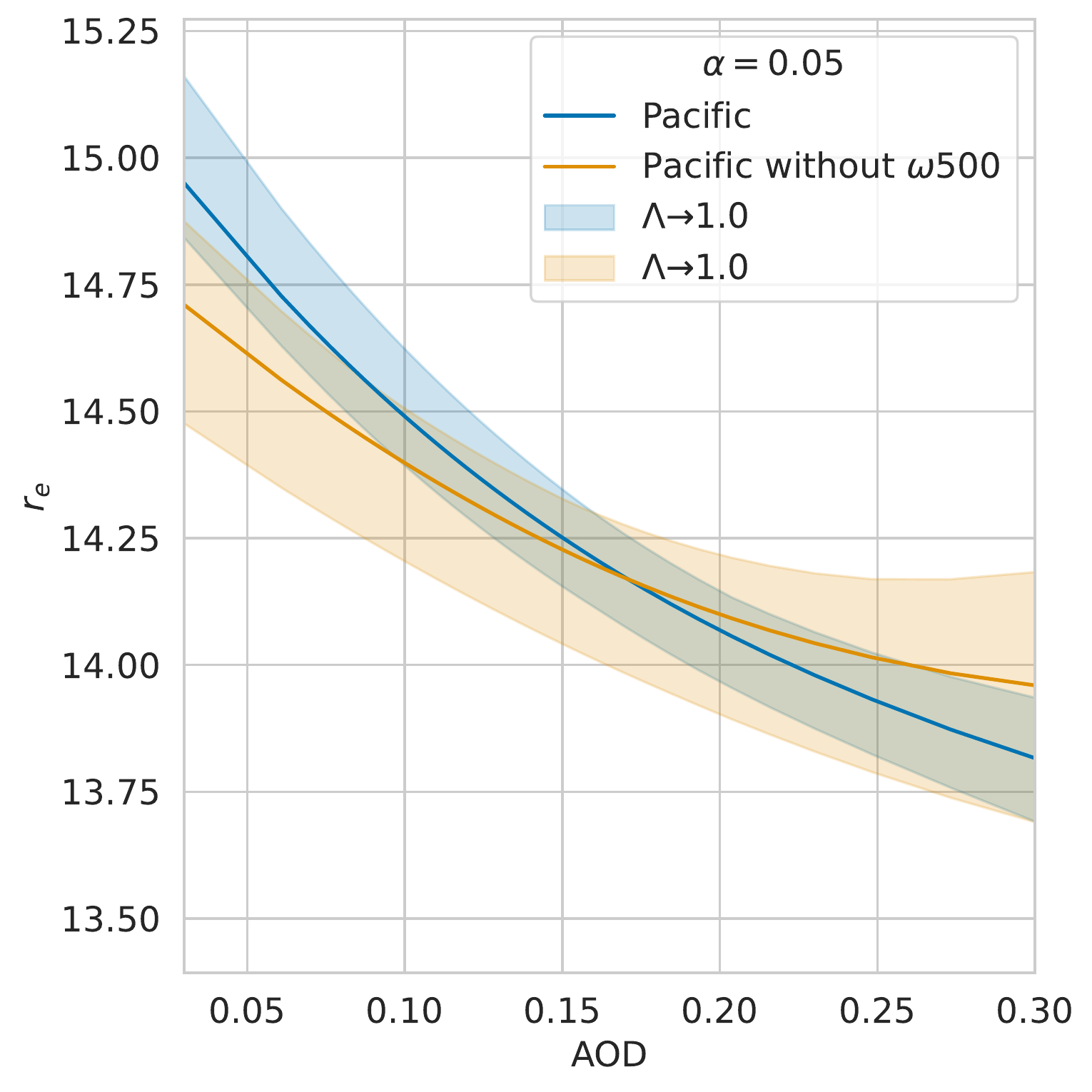}
        \label{subfig:apo_tr_lrp-vs-lrpw500}
        \caption{Unscaled}
    \end{subfigure}
    \begin{subfigure}[b]{0.46\textwidth} 
        \includegraphics[width = \textwidth]{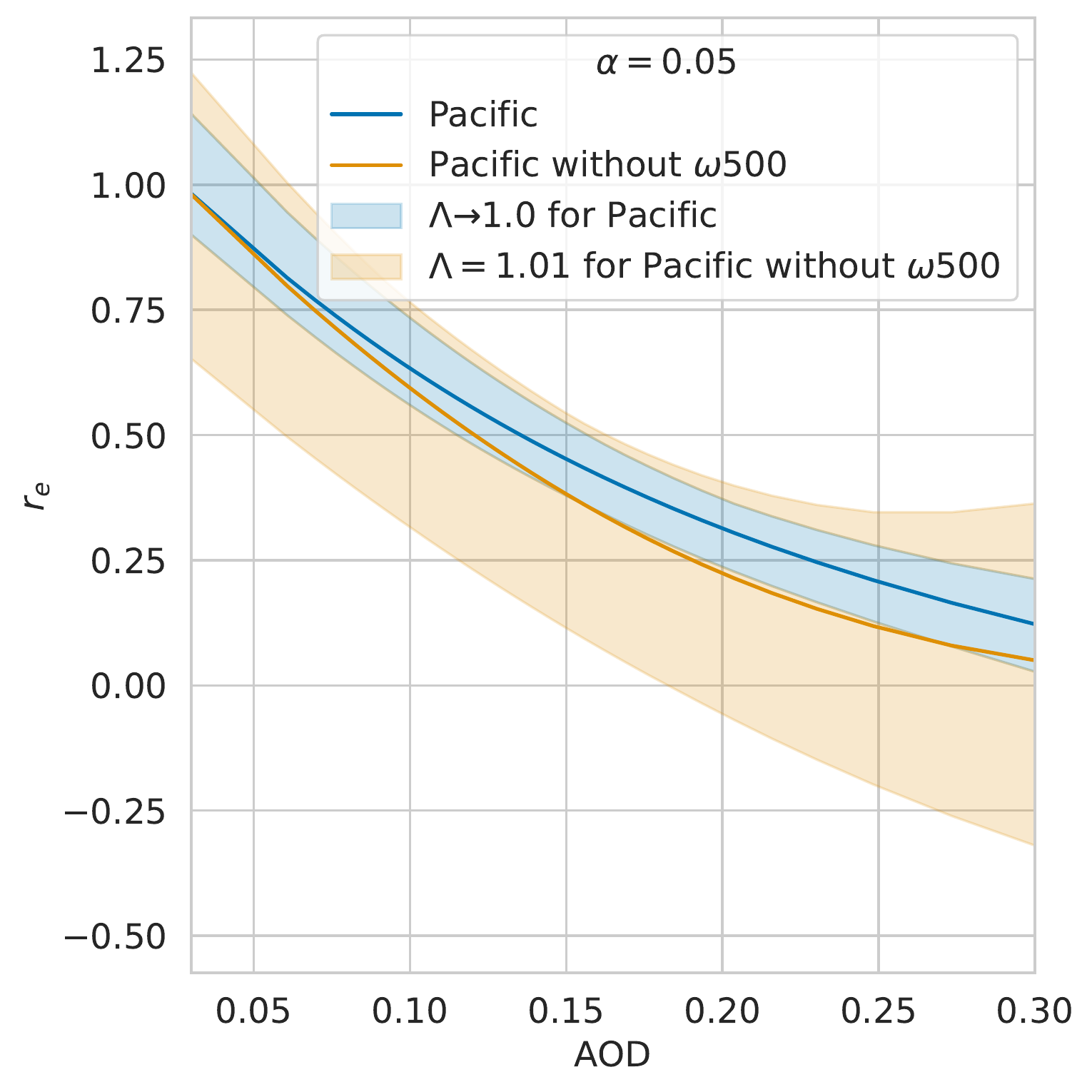}
        \label{subfig:apo-scaled_tr_lrp-vs-lrpw500}
        \caption{Scaled by min-max with appropriate $\Lambda$}
    \end{subfigure}
    \caption{Dose-response curves with or without vertical motion at 500 mb ($\omega500$) as a covariate.}
    \label{fig:apo_tr_lrp-vs-lrpw500}
\end{figure}

In \Cref{fig:apo_tr_lrp-vs-lrpw500} we compare the same region with different covariates to identify an appropriate $\Lambda$. 
We fit one model on data from the Pacific (blue) and one model from the Pacific omitting $\omega500$ from the covariates (orange). 
The shaded bounds in blue are the ignorance region for $\Lambda \to 1$ for the Pacific. 
We then find the $\Lambda$ that results in an ignorance interval around the Pacific omitting $\omega500$ that covers the Pacific model prediction. 
From this, we can infer how the parameter $\Lambda$ relates to the inclusion of covariates in the model. We show that we need to set $\Lambda=1.01$ to account for the fact that we are omitting $\omega500$ from our list of covariates.
We also note that the slopes of the dose-response curves are slightly different, with worse predictions when omitting $\omega500$ from the covariates, as expected.

This work attempts to set a new methodology for setting $\Lambda$ which can be summarised as followed.
Working with two datasets, which vary in only aspect, we train two different models: (i), the control model, and (ii), the experimental model. 
After training both models, we plot the dose-response curves for (i) and (ii) on the same plot. 
We can compare the shape and slope of these curves as well as their uncertainty bounds under the unconfoundedness assumption by plotting the ignorance region for $\Lambda \to 1$ for both models. 
Then, we are interested in setting $\Lambda$ for model (ii) such that the uncertainty bounds cover the entire ignorance region of model (i) under the unconfoundedness assumption. 
For this, we are interested in comparing the slopes and thus min-max scale both curves.

\end{document}